\documentclass[journal]{IEEEtran}
\ifCLASSINFOpdf
   \usepackage[pdftex]{graphicx}
   \graphicspath{{./}{./images/}{./figures/}}
\else
\fi
\newcommand*\diff{\mathop{}\!\mathrm{d}}
\usepackage[utf8]{inputenc}
\usepackage[T1]{fontenc}
\usepackage{amsmath,color}
\usepackage{amsfonts}
\usepackage{multirow}

\usepackage{xifthen}
\newcommand{\comment}[2][]{{\color{red}
		\ifthenelse{\isempty{#1}}%
		{}
		{\textit{#1}:~}
		#2}}
	
\usepackage{tabularx}
\newcolumntype{L}[1]{>{\hsize=#1\hsize\raggedright\arraybackslash}X}%
\newcolumntype{R}[1]{>{\hsize=#1\hsize\raggedleft\arraybackslash}X}%
\newcolumntype{C}[1]{>{\hsize=#1\hsize\centering\arraybackslash}X}%
\usepackage{xcolor}
\usepackage[colorlinks]{hyperref}
\AtBeginDocument{%
	\hypersetup{
		colorlinks=true,
		allcolors=purple,
}}

\begin{document}

\title{Model predictive altitude and velocity control in ergodic potential field directed multi-UAV search}

\author{Luka~Lanča, Karlo~Jakac and Stefan~Ivić*
\thanks{L.~Lanča, K.~Jakac and S.~Ivić are with Faculty of Engineering, University of Rijeka, Rijeka, Croatia. *Corresponding author, e-mail: {stefan.ivic@riteh.hr}}
}

%
%

\markboth{}%
{}
%



\maketitle

\begin{abstract}
	This research addresses the challenge of executing multi-UAV survey missions over diverse terrains characterized by varying elevations. The approach integrates advanced two-dimensional ergodic search technique with model predictive control of UAV altitude and velocity. Optimization of altitude and velocity is performed along anticipated UAV ground routes, considering multiple objectives and constraints. This yields a flight regimen tailored to the terrain, as well as the motion and sensing characteristics of the UAVs. The proposed UAV motion control strategy is assessed through simulations of realistic search missions and actual terrain models. Results demonstrate the successful integration of model predictive altitude and velocity control with a two-dimensional potential field-guided ergodic search. Adjusting UAV altitudes to near-ideal levels facilitates the utilization of sensing ranges, thereby enhancing the effectiveness of the search. Furthermore, the control algorithm is capable of real-time computation, encouraging its practical application in real-world scenarios.
\end{abstract}


\section{Introduction}

The main objective of Search and Rescue (SAR) operations is to locate lost, missing, or injured individuals as quickly as possible, to minimize their suffering and reduce the risk of environmental effects, injuries, and other hazards. These operations often occur in remote or inaccessible areas, such as wilderness, urban ruins, or the ocean, which makes them particularly challenging. Autonomous UAVs can offer numerous benefits in SAR operations, as they can efficiently and quickly survey a large area at a relatively low cost, which is particularly important in situations where time is of the essence.

While conducting area surveys, keeping the UAV within a designated altitude range is crucial to achieve good ratio between area coverage and image detail. This becomes especially challenging over uneven terrain. The primary objective during the search is speed, often implying the use of the highest possible velocity. Maximizing velocity while ensuring a collision free path, adhering to desired altitude goals and respecting the UAV's technical constraints and capabilities poses a significant challenge. Employing multiple UAVs can further expedite the search process. However, real-time control of multiple UAVs requires high computational efficiency, which adds additional complexity to the problem considering the need for altitude and velocity control while simultaneously avoiding collisions and restricted zones.

To tackle the outlined challenges, we employed the Heat Equation Driven Area Coverage (HEDAC) algorithm, designed for two-dimensional area search supporting the use of multiple UAVs. It was coupled with Model Predictive Control (MPC) optimization, performing velocity and altitude control, which enabled flight within a three-dimensional domain. UAV collision avoidance in regard to the boundary and other UAVs is managed within a two-dimensional domain, automatically preventing flyovers. Area coverage is applied utilizing a sensor with pyramidal-shaped Field Of View (FOV) that simulates an orthophoto camera. This sensor is integrated with a sensing function that mirrors the accuracy of the employed detection model. Terrain is modeled using open access Digital Elevation Models (DEM) which have a resolution of 30 meters throughout most areas. A minimum UAV altitude requirement of 30 meters is assigned, accounting for potential terrain model inaccuracies, added height from trees and vegetation, human-made structures like sheds or electricity transmission lines, along with an additional safety clearance. The proposed method is validated using three test cases of varying terrain complexity, showcasing different search scenarios, on which we performed survey simulations and evaluated UAVs' performance and overall search success.

\section{Literature overview }

Several research papers have contributed to the advancement of UAVs and their use in numerous applications. In the context of SAR, they can work in independently or conjunction with a team of ground searches as described in \cite{goodrich2008supporting}. In this case, camera-equipped UAVs are used to find clues and guide the search team towards the missing persons. UAVs with pre-programmed custom missions in \cite{silvagni2017multipurpose} are utilized for a mountain avalanche SAR scenario. They incorporate sensors that continuously update the current mission based on environmental readings. The authors in \cite{lin2009uav} describe 2D path planning methods that utilize straight paths and 90-degree turns to effectively cover the probability distribution which is employed in Wilderness Search And Rescue (WiSAR) application. 

Swarm-based or multi-agent approaches are frequently employed to address coverage problems, exemplified by the work in \cite{atincc2020swarm}, which introduces a swarm-based coverage method utilizing groups of cooperating agents guided by a leader. Another example is provided by \cite{arnold2018search} which simulates the usage of UAV swarms in aiding SAR missions during natural disasters such as earthquakes or tsunamis. It also incorporates considerations such as battery threshold which, when reached, forces the UAV to break from search formation and continue navigating towards the charging station. In a related study, the authors in \cite{song2022multi} explore multi-agent solutions for coverage of disaster areas with agent endurance constraints. In order to bring the energy consumption to a minimum, \cite{gao2022energy} uses mid-field game approach to control velocities of a large number of UAVs. Research in \cite{alotaibi2019lsar} presents a search algorithm which decomposes the domain into layers and executes the search using multiple cooperating UAVs. 

Coverage task with imposed altitude goals is presented in \cite{garrido2023fast} which focuses on pre-defined agent trajectories computed using the fast marching method with additional terrain-following functionality. The inspection is performed by multiple agents, with one leader and several followers that maintain a formation during the search. Another approach for generating terrain-following trajectories is presented in \cite{kosari2015optimal} which employs neural networks to generate two-dimensional trajectories that meet constraints such as drone dynamics, minimum and maximum flight height.

Implementation and experimental comparison of different motion planners for area coverage problem using real fixed-wing UAV is presented in \cite{xu2011optimal}. It uses robot-specific controller which takes the optimal coverage path and makes the motion plan that satisfies the dynamic constraints of the aircraft. It also provides comparison between simulation results and real-world experiments. Another practical implementation is presented in \cite{li2018use} where a single multi-rotor UAV equipped with appropriate sensors is used on a field test for radioactive source search.

Optimization techniques such as Particle Swarm Optimization (PSO) are often used for trajectory generation as presented in \cite{phung2021safety} where authors compare different variations of PSO for computing three-dimensional trajectories that avoid obstacles and satisfy altitude constraints. Generated trajectories are validated in field test experiments, demonstrating their validity for real-world applications. Additionally, study in \cite{sun2011intelligent} presents an intelligent flight task algorithm that employs PSO, skeletonization, and B-spline curves to determine optimal UAV flight routes in complex topographies. The algorithm utilizes skeletonization to reduce complexity and PSO to determine the best control points for a smooth B-spline curve flight route.

Past achievements, current developments, and future research regarding MPC are presented in \cite{mayne2014model}. It is also known as Receding Horizon Control (RHC) and is a commonly used technique for trajectory planning as demonstrated by \cite{kuwata2005robust} and \cite{bellingham2002receding}. In a study carried out in \cite{ahmadzadeh2007cooperative}, RHC was employed to address the cooperative coverage problem while considering collision avoidance constraints. The approach utilized centralized solving and focused on the use of heterogeneous autonomous vehicles operating at fixed altitudes. They were subject to constant forward velocities and minimal turning radius constraints. In \cite{romero2022model}, model predictive contouring control method was utilized to perform time-optimal quadrotor flight through multiple waypoints in a closed loop. It was compared to standard MPC and expert human pilots in a real-world experiment. 

In \cite{stastny2015collision} MPC and an artificial potential field are used to navigate fixed-wing UAVs while avoiding collisions by utilizing points of repulsive potential. Collision avoidance method for UAVs traveling at constant velocity is presented in \cite{marchidan2020collision} which generates local guidance vector fields to evade static and moving obstacles. Decentralized, asynchronous 3D trajectory planner for multi-agent systems, generating collision free routes in an environment with static and moving obstacles was presented in \cite{tordesillas2021mader}.

Real-time trajectory generation system for flight in complex environments is presented in \cite{gao2017gradient} where a quadrotor equipped with onboard sensors collects environment data and utilizes gradient information to produce smooth, collision-free trajectory through optimization. Further improvements in efficiency and convergence rate are demonstrated in \cite{zhou2019robust} where the proposed method is validated through real-world experiments. Method for collaborative autonomous exploration using decentralized multi-UAV system was presented in \cite{zhou2023racer}. Each UAV navigates and explores the designated space, generating a volumetric map by collecting environment data. It exchanges map information with nearby UAVs when communication is available. The proposed method is evaluated in simulated and real-world scenarios.

The foundation of this paper is the Heat Equation Driven Area Coverage (HEDAC) algorithm which was initially introduced in \cite{ivic2016ergodicity} and has since been further improved for search applications incorporating agent sensing and detection \cite{ivic2020motion}. To enhance its capabilities, the Finite Element Method (FEM) was utilized for solving underlying heat equation in \cite{ivic2022constrained}. This integration enabled greater control over the agents' motion, facilitating the handling of irregularly shaped domains and inter-domain obstacles without incurring additional computational costs. The versatility of the algorithm has led to its application in various contexts. For instance, it was adopted for multi-agent maze exploration \cite{crnkovic2023fast} and it can be utilized for controlling multiple UAVs performing non-uniform crop spraying \cite{ivic2019autonomous}. In \cite{bilaloglu2023whole}, HEDAC was employed in a whole-body ergodic exploration method for robotic manipulators. In \cite{zheng2022distributed}, it was used to solve a coverage problem by dividing the search region into multiple sub-regions. Additionally, in \cite{low2022drozbot}, the algorithm was employed to create artistic portraits by reformulating the coverage problem.

These papers collectively contribute to the advancement of UAV flight control, trajectory planning and SAR missions, offering insights into different control algorithms, obstacle avoidance methods, coverage strategies, and collaborative approaches.

\section{UAV motion model}
\label{sec:motion_model}

In order to model the motion of multiple UAVs within a three-dimensional region $\Omega_{3D} \in \mathbb{R}^3$, where each of the total $n$ UAVs is identified by the index $i$, we consider the the following control variables regulated in time $t$:
\begin{itemize}
	\item velocity intensity $\rho_i(t) \in [0, 1]$,
	\item incline angle $\varphi_i(t) \in [\varphi_{min, i}, \varphi_{max, i}]$, and
	\item yaw angular velocity $\omega_i(t) \in [-\omega_{max, i}, \omega_{max, i}]$.
\end{itemize}

The incline parameter $\varphi$ denotes the angle between the velocity vector, which is tangential to the UAV trajectory, and the horizontal plane. It is important to highlight, especially in the context of multi-rotor UAVs, that the incline angle $\varphi$, representing the slope of the resulting trajectory, is distinct from the aircraft pitch. Aircraft pitch refers to the angle between the longitudinal axis of the aircraft and the horizontal plane. Additionally, we do not take into account possible lateral motion resulting from adjustments in multi-rotor roll.

To authentically incorporate the distinct difference in velocity characteristics UAVs exhibit when flying horizontally, ascending, or descending, we utilize the limit velocity function $v(\varphi)$. It represents the maximal absolute velocity the UAV can achieve for a given incline. The function provides a concise description of the technical (velocity) characteristics of the UAV and can be determined experimentally for a specific aircraft. 

\begin{figure*}[!htb]
	\centering
	\includegraphics[width=0.9\linewidth]{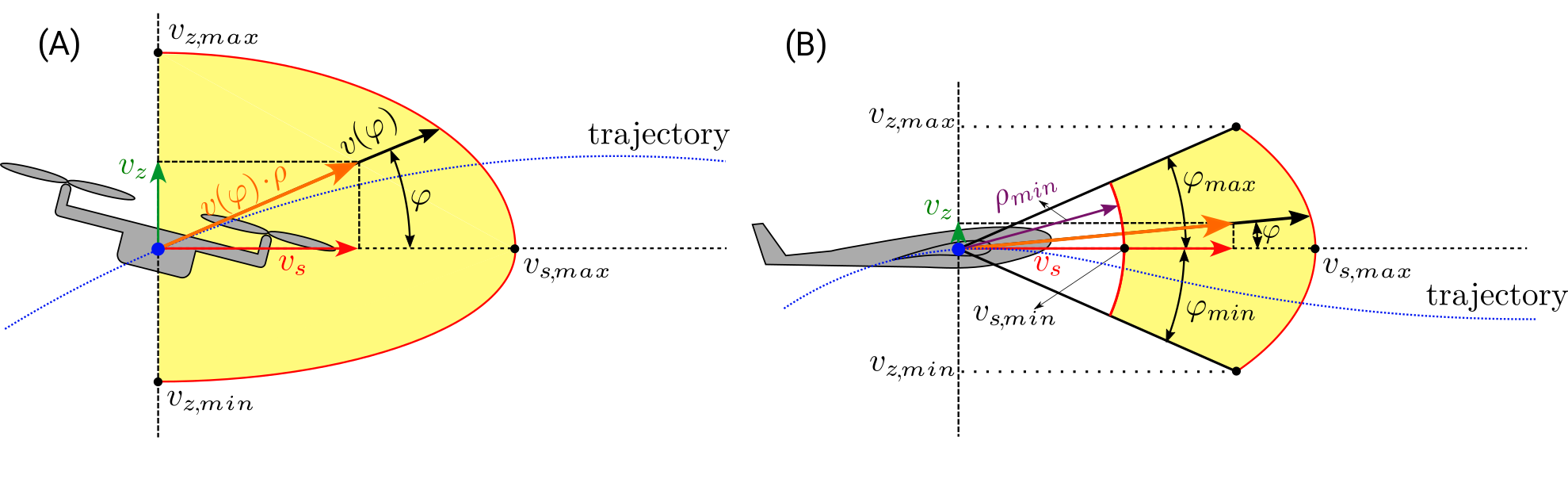}
	\caption{ Definition of velocity components and constraints for multi-rotor (A) and fixed-wing (B) UAVs. Vertical and horizontal velocity components are depicted by the red and green vectors, respectively, while the orange vector represents the total velocity. The red line illustrates the velocity limits corresponding to various inclines and shaded yellow area represents the feasible region for the tip of the total velocity vector, originating from the center of aircraft's mass. 
		The feasible region for multi-rotor UAV allowing different dynamics of ascending and descending motions, hence it is limited with two elliptical arcs defined with $v_{s,max}$, $v_{z,min,}$ and $v_{z,max,}$ characteristics of the aircraft. The motion of fixed-wing UAV is limited with a circular and an elliptical arc constructed using measurable UAV parameters $v_{s,min}$ (providing $\rho_{min}$ limit), and $v_{s,max}$, $v_{z,min}=-v_{z,max,}$ and $\varphi_{min}=-\varphi_{max}$, respectively. From the depicted image, it is evident that the fixed-wing UAV's minimum horizontal velocity can fall below the prescribed value of $v_{s,min}$. However, this discrepancy is disregarded due to the defined limits of $\varphi$, which ensure that the aircraft cannot deviate more than 4\% below its minimum horizontal velocity.} 
	\label{fig:incline-vs-pitch}
\end{figure*}

For simplicity, we approximate the limit velocity function with an (asymmetric) ellipse constructed using 3 distinctive velocities: maximal horizontal velocity $v_{s,max} \equiv v(\varphi = 0)$, maximal ascending velocity $v_{z,max} \equiv v(\varphi_{max})$ and maximal descending velocity $v_{z,min} \equiv v(\varphi_{min})$ as shown in Figure \ref{fig:incline-vs-pitch}. 

To accurately simulate the movement of various types of UAVs, we introduce additional constraints. When modeling fixed wing UAVs we need to limit the range of $\varphi$ values to avoid phenomena such as aircraft stall. This effect usually happens when wings reach an angle of attack around $15^\circ \approx 0.26 \text{ rad}$, therefore we impose $\varphi$ limits in $[-0.25, 0.25]$. For multi-rotor UAVs, we limit the $\varphi$ parameter from $\varphi_{min} = - \pi / 2$ which represents vertical descent and $\varphi_{max} = \pi / 2$ that corresponds to vertical ascent. By defining $\varphi$ in this manner, multi-rotor UAVs gain the capability to execute purely vertical motion, without any horizontal movement.

The velocity intensity $\rho$ indicates the percentage of the limit velocity that is actually utilized. In contrast to multi-rotor UAVs, fixed-wing UAVs inherently require a certain amount movement to generate lift. Consequently, we introduce a minimum horizontal velocity, denoted as $v_{s,min}$, to effectively restrict $\rho$ within the range $\left[\rho_{min}, \rho_{max} \right] = \left[ v_{s,min}/v_{s,max}, 1 \right]$. Horizontal and vertical velocities can now be defined as:
\begin{equation}
	v_{s,i}(\rho_i, \varphi_i) = \rho_i \cdot v_i(\varphi_i) \cdot \cos(\varphi_i)
	\label{eq:v_s}
\end{equation}
\begin{equation}
	v_{z,i}(\rho_i, \varphi_i)= \rho_i \cdot v_i(\varphi_i) \cdot \sin(\varphi_i).
	\label{eq:v_z}
\end{equation}

Now we can define the UAV motion, i.e. the trajectory $\mathbf{X}_i(t)=\left[ x_i(t), y_i(t), z_i(t) \right] \in \Omega_{3D}$, that is governed by flight control parameters $\omega_i(t)$, $\rho_i(t)$, $\varphi_i(t)$:
\begin{equation*}
	\frac{\diff x_i}{\diff t} = v_{s,i}(\rho_i, \varphi_i) \cdot \cos\theta_i
\end{equation*}
\begin{equation*}
	\frac{\diff y_i}{\diff t} = v_{s,i}(\rho_i, \varphi_i) \cdot \sin\theta_i
\end{equation*}
\begin{equation*}
	\frac{\diff z_i}{\diff t} = v_{z,i}(\rho_i, \varphi_i)
\end{equation*}
where $\theta$ is the heading angle regulated by the yaw angular velocity $\omega$:
\begin{equation*}
	\frac{\diff \theta_i}{\diff t} = \omega_i(t).
\end{equation*}

The constraints for the yaw angular velocity are defined later in section \ref{sec:horizontal_search}.

\section{Sensing model and search evaluation}

Though UAVs operate in three-dimensional space, their primary task is to explore and observe the terrain surface which can be considered as a problem of two-dimensional ergodic search in the horizontal domain $\Omega_{2D}$. The terrain is defined using a terrain height function $z_T: \Omega_{2D} \mapsto \mathbb{R}$. In order to define the sensing model we assume that the UAV trajectory and initial probability distribution $m_0: \Omega_{2D} \mapsto \mathbb{R}$ is known. The initial probability distribution is an assessment of spatial probability of targets (or missing persons) on a terrain surface projected to the horizontal plane $\Omega_{2D}$ at time $t=0$. 

In order to simulate the surveillance performance of the UAV, the geometric (field of view) and sensing (detection probability) properties of the camera need to be considered. Both of this aspects are incorporated in the instantaneous detection probability function defined in UAVs local coordinates $\in \mathbf{R}^3$:
\begin{equation*}
	\psi_i(\mathbf{R}) = 
	\begin{cases}
		\Gamma_i(\|\mathbf{R}\|) & \text{ if } \mathbf{R} \in \Omega_{FOV,i} \\
		0 & \text{ otherwise.} 
	\end{cases}
\end{equation*}

\begin{figure}[!htb]
	\centering
	\includegraphics[width=1\linewidth]{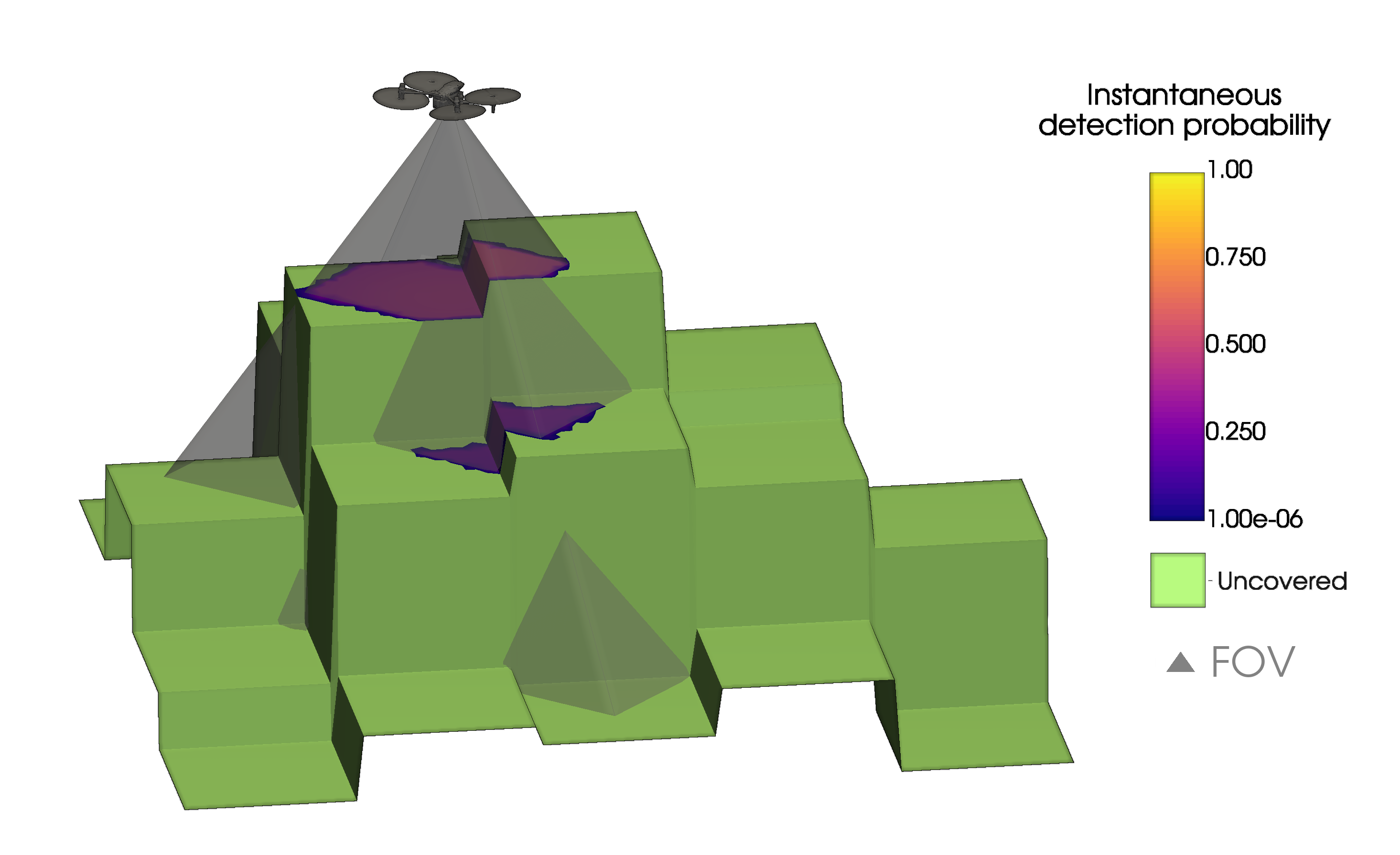}
	\caption{A single UAV utilizes the instantaneous detection probability on the terrain surface within its FOV. The detection probability is determined by the distance from the camera sensor and is visually depicted as a gradient spanning from purple to yellow. The unobserved area is depicted in green, while the sensor's FOV is illustrated using a semi-transparent gray pyramid. It is important to note that only the areas within the FOV pyramid that have successfully passed the ray trace check are affected by the sensor, not the entire enclosed space.} 
	\label{fig:sensing_fun}
\end{figure}

Surveillance is performed by a camera sensor that produces a rectangular image that envelops terrain within pyramidal FOV $\Omega_{FOV} \in \mathbb{R}$. Dependent on the UAV heading direction, FOV is defined with angles $\gamma_1$ and $\gamma_2$ between two lateral and two longitudinal sides of the pyramid, respectively. Since we are considering uneven terrain and different flight altitudes, the recorded image contains the terrain surface encapsulated by an rectangular pyramid as shown in Figure \ref{fig:sensing_fun}. The apex of the pyramid coincides with the camera's position that is located at $\mathbf{X}_i$ and it is the origin for local coordinates $\mathbf{R}$. The detection probability is set to 0 for any point outside the sensor's FOV, but also for points outside the line of sight which is determined with a ray trace check. For the area within the sensor's scope, detection probability $\Gamma$ is calculated depending on $\|\mathbf{R}\|$ which equals the distance between the sensor and the observed point.

To determine whether an arbitrary point $\mathbf{p} \in \Omega_{2D}$ is sensed by the sensor at $\mathbf{X}$ we need a transformation to local coordinates that includes three-dimensional translation and a rotation in horizontal plane depending on the UAV heading direction $\theta$:

\begin{equation*}
	\mathbf{R}
	=  
	\begin{bmatrix}
		\cos\theta_i & -\sin\theta_i & 0\\
		\sin\theta_i & \cos\theta_i & 0\\
		0 & 0 & 1\\
	\end{bmatrix} 
	\left(\mathbf{X}_i - \left[\mathbf{p}_x, \mathbf{p}_y, z_T(\mathbf{p})\right]^T\right).
\end{equation*}
Note that $\mathbf{p}$ is projected on the terrain surface using height coordinate $z_T(\mathbf{p})$.

The overall detection effect of all $n$ UAVs involved in the search is expressed with coverage density which evaluates detection probability along their paths:
\begin{equation*}
	c(\mathbf{p}, t) = \sum_{i = 1}^{n} \int_{0}^{t} \psi_i(\mathbf{R}(\mathbf{X_i}(t), \mathbf{p}))dt.
\end{equation*}

The probability of undetected target presence at any point $\mathbf{p}$ and time $t$ is determined by combining initial target presence probability and conducted sensing after time $t$ at the same location:

\begin{equation*}
	m(\mathbf{p}, t) = m_0(\mathbf{p}) \cdot e^{-c\left(\mathbf{p}, t \right) }.
\end{equation*}

To evaluate the overall search success we integrate $m$ to obtain the survey accomplishment on the entire domain:
\begin{equation*}
	\eta(t) = 1 - \int_{\Omega_{2D}} m \left(\mathbf{p}, t \right) d\mathbf{p}.
\end{equation*}

The flight control system needs to be designed to dynamically adjust parameters $\omega(t)$, $\rho(t)$, and $\varphi(t)$ with the objective of minimizing the value of $\eta(t)$.

\section{Horizontal search control} \label{sec:horizontal_search}

The three-dimensional trajectory of the UAV is governed by the combined influence of horizontal and vertical movements, allowing us to calculate them separately. We begin by computing the UAV's horizontal path, which represents its movement within the search domain $\Omega_{2D}$. Following that, we perform altitude and velocity optimization to determine the final trajectory and complete control of UAVs.

Effective horizontal movement control is crucial for a successful search, as it directs UAVs to the points of maximum target potential. Altitude control enables horizontal movement because it guarantees compliance with the minimal altitude value along uneven terrain while simultaneously striving for optimal altitude which ensures adequate area coverage and detection rate. Velocity control guarantees the feasibility of executing the 3D path generated combining horizontal and vertical path requirements considering UAV motion constraints.

In accordance with the work in \cite{ivic2020motion} we calculate the potential field $u(\mathbf{p}, t)$, which essentially guides the UAVs to the zones with the highest probability of undetected target presence, by solving the partial differential equation 
\begin{equation*}
	\alpha \cdot \Delta u (\mathbf{p}, t) = \beta \cdot u (\mathbf{p}, t) - m(\mathbf{p}, t)
\end{equation*}
using the boundary condition 
\begin{equation*}
	\frac{\partial u}{\partial \textbf{n}} = 0
\end{equation*}
where \textbf{n} denotes the outward normal to the domain boundary $\partial \Omega_{2D}$. HEDAC parameters $\alpha > 0$ and $\beta > 0$ can be adjusted to achieve different search behavior. $\alpha$ governs the smoothness of the probability field and therefore dictates whether the search focus is local or global while $\beta$ serves as a numerical stability factor and has a weaker effect on the search performance.  

To ascertain the travel direction we introduce the vector field 
\begin{equation*}
	\mathbf{u}(\mathbf{p},t) = \frac{\nabla u(\mathbf{p},t)}{\left| \nabla u(\mathbf{p},t) \right| }.
\end{equation*}
which equals the normalized gradient of the potential field $u$.

The angle between the current direction vector $\mathbf{v}_i(t)$ and vector $\mathbf{u}(x_i, y_i, t)$ provides a change of heading we want to realize in the control step $\Delta t$. To alter the heading, the UAV pivots around its yaw axis, utilizing yaw angular velocity

\begin{equation*}
	\omega_{i} = \frac{1}{\Delta t}\text{arccos} \left( \frac{\mathbf{v}_i(t) \cdot \mathbf{u}(x_i, y_i, t)} {\left| \mathbf{v}_i(t) \right| } \right).
\end{equation*}

Each UAV is characterized by maximal achievable angular velocity denoted as $\omega_{lim}$, representing its technical limitation. Therefore, we require that $|\omega| \leq \omega_{lim}$. 

Horizontal velocity is determined by testing various velocity values along a specific path segment (described later in section \ref{sec:trial_trajectory}). To guarantee the feasibility of movement along the path regardless of the velocity used, we enforce a path curvature constraint defined by the minimal turning radius $R_{min}$. The highest $\omega$ value is attained when moving at maximum horizontal velocity along a path characterized by a curvature of $R_{min}$.

Considering all mentioned requirements for $\omega$, we define maximal omega value $\omega_{max}$ and impose a constraint 

\begin{equation}
	\left| \omega \right| \leq \omega_{max} = \text{min} \left( \omega_{lim}, \frac{v_{s,max}}{R_{min}} \right). 
	\label{eq:omega_constarints}
\end{equation}

\subsection{Collision avoidance}

\begin{figure*}[h!]
	\centering
	\includegraphics[width=0.8\linewidth]{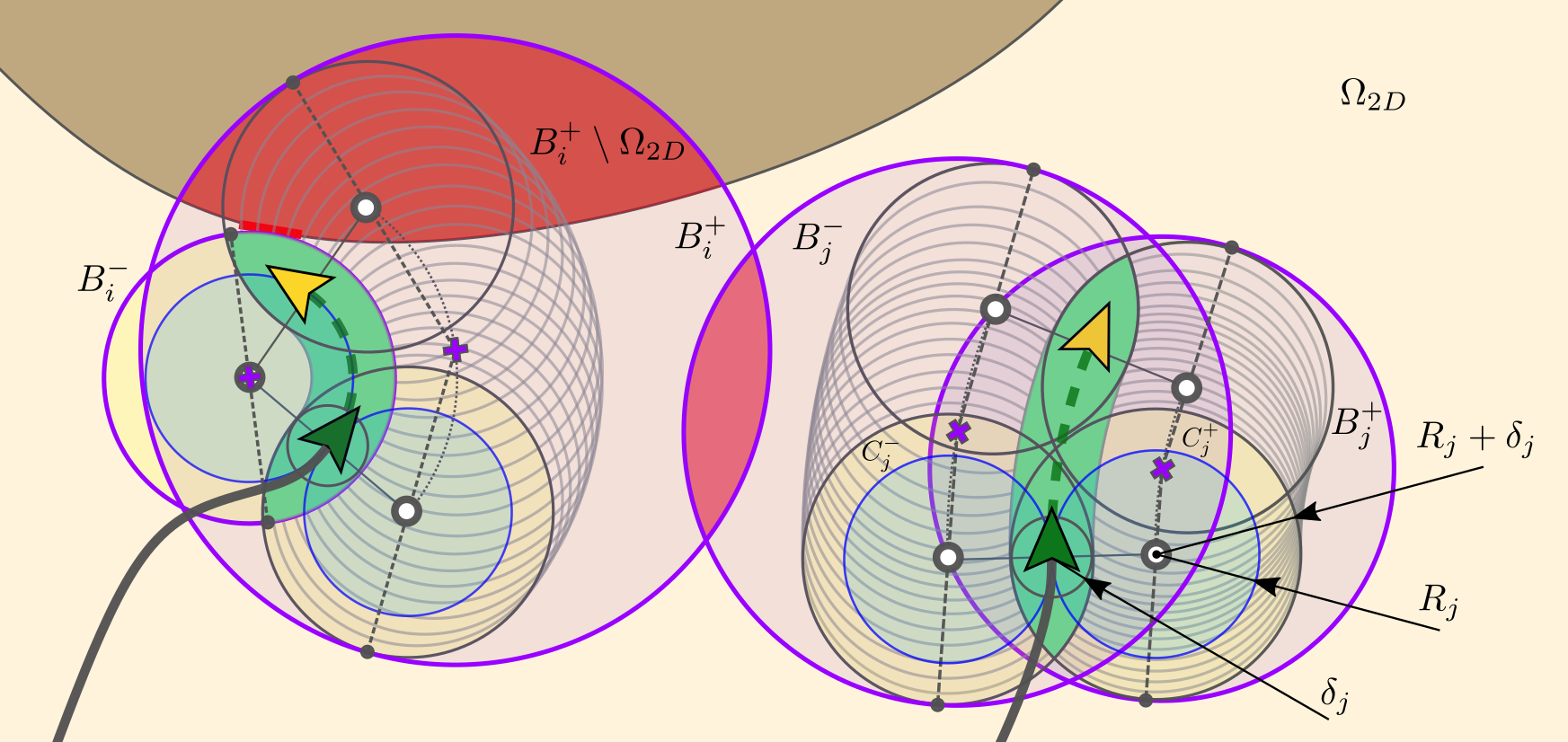}
	\caption{Two UAVs performing a collision check using bounding circles $B^+$ and $B^-$ shown in purple. The current aircraft position is indicated by the green arrow, while the yellow arrow represents the next position for $v_s = v_{s,max}$ and $\omega$ that is currently being tested for collision. The blue circles depict radical "escape routes" that are executed using $\pm \omega_{max}$, while the yellow shaded circles represent previously used clearing circles $C^+$ and $C^-$. The gray circles illustrate a subset of potential clearing circle positions corresponding to various horizontal velocities. The clearing circles for $v_s = 0$ (current position) and $v_s = v_{s,max}$ are internally tangent to the bounding circle which shares its center (shown as purple "+") with the clearing circle for $v_s = v_{s,max}/2$. The area shaded in green should always be collision free to satisfy minimum clearance $\delta$ constraint.} 
	\label{fig:collision_circles}
\end{figure*} 

\label{subsec:hedac_collison_avoidnace}
The final step is to check and adapt $\omega$ values to ensure a collision free path. The procedure is analogue to previous work \cite{ivic2022constrained} with the only difference being that the collision is checked with bounding circles $B^+$ and $B^-$ instead of clearing circles $C^+$ and $C^-$ which were previously used. Since UAV velocities are variable and unknown at this stage, the positions of the clearing circles can differ based on different horizontal velocities. Consequently, it is necessary to examine collisions for all potential locations of the clearing circles, therefore we use the bounding circle which contains all possible clearing circles as shown in Figure \ref{fig:collision_circles}. In summary, we firstly check all UAVs for collision using their initial $\omega_i$. Only if there is a collision, $\omega_i$ values are optimized in range $\left[ -\omega_{max,i}, \omega_{max,i}\right]$. Bounds of the optimization range represent possible escape omega $\omega_{esc,i}$ values which are utilized during the collision avoidance (or escape) maneuver while the path executed using one of the bounds has to be collision free at any given moment. The optimization goal is to acquire $\omega_i$ values that are closest to the initial $\omega_i$ values (computed from $\mathbf{v}_i(t)$ and $\mathbf{u}(x_i, y_i, t)$) but do not produce collision.

After the final $\omega_i$ values are attained, we analyze whether the right or left side of the UAV is available for the escape maneuver (whether $B^+$ or $B^-$ is collision free) and determine the escape omega $\omega_{esc,i}$ values. Escape omega can be either $-\omega_{max,i}$ or  $\omega_{max,i}$, so we choose the value closer to final $\omega_i$ that yields a collision free route. 

\section{Velocity and altitude control} 

Velocity and altitude control are achieved using MPC technique which determines $\rho$ and $\varphi$ regimes that produce an optimal predicted flight for time window $[t, t+\tau_{max}]$. By employing the horizontal control method discussed in the preceding section, we generate a two-dimensional (projected to horizontal plane) predicted path and acquire the terrain's elevation profile along this trajectory. To achieve predefined goals and adhere to constraints, an optimal flight path is established by adjusting the $\rho$ and $\varphi$ regimes within the time window of $[t, t+\tau_{max}]$.

\subsection{Trial trajectories and trial control functions}
\label{sec:trial_trajectory}

First, we introduce a predicted path in horizontal plane, achievable using maximum velocity ($\rho=1$) and without vertical movement ($\varphi=0$):
\begin{equation*}
	\frac{\diff\bar{x}_i}{\diff\tau} = v_{s,max,i} \cdot \cos\bar{\theta}_i
\end{equation*}
\begin{equation*}
	\frac{\diff\bar{y}_i}{\diff\tau} = v_{s,max,i} \cdot \sin\bar{\theta}_i
\end{equation*}
\begin{equation*}
	\frac{\diff\bar{\theta}_i}{\diff\tau} = \bar{\omega}_i.
\end{equation*}
The UAV trial yaw angular velocity is calculated from unchanging (in time window $[0, \tau_{max}]$) potential field $u$ obtained at time $t$:
\begin{equation*}
	\bar{\omega}_{i}(\tau) = \frac{1}{\Delta t}\arccos \left( \frac{\bar{\mathbf{v}}_i(\tau) \cdot \mathbf{u}(\bar{x}_i(\tau), \bar{y}_i(\tau), t)} {\left| \bar{\mathbf{v}}_i(\tau) \right| } \right).
\end{equation*}
The calculated yaw is verified to ensures constraint \eqref{eq:omega_constarints} and collision avoidance for $[0, \Delta t]$. 

For convenience, a path length function $\bar{s}(\tau)$ can be easily obtained from
\begin{equation*}
	\frac{\diff \bar{s}}{\diff \tau} = v_{s,max,i}
\end{equation*}
as $\bar{s}=v_{s,max}\cdot\tau$ which allows for the trajectory parametrization $\bar{x}_i(s)$ and $\bar{y}_i(s)$.

According to $\tilde{\rho}$ and $\tilde{\varphi}$, we define trial trajectory length function $\tilde{s}$ from
\begin{equation*}
	\frac{\diff \tilde{s}_i}{\diff \tau} = v_{s,i}\left(\tilde{\rho}_i(\tau), \tilde{\varphi}_i(\tau)\right)
\end{equation*}
and utilize it in order to determine a trial trajectory in time window $[0, \tau_{max}]$ using above mentioned length parametrization of $\bar{x}$ and $\bar{y}$:
\begin{equation*}
	\tilde{x}_i(\tau) = \bar{x}_i\left(\tilde{s}_i(\tau)\right)
\end{equation*}
\begin{equation*}
	\tilde{y}_i(\tau) = \bar{y}_i\left(\tilde{s}_i(\tau)\right).
\end{equation*}
Note that UAV trial trajectory, in general, does not pass entire available predicted path in $[0, \tau_{max}]$ due to possible partial use of velocity intensity ($\rho(\tau) < 1$) or possible ascending or descending trial flight regimes ($\varphi(\tau)\neq 0$).
The trial yaw angular velocity $\tilde{\omega}$ can be calculated using the parametrization $\bar{\omega}_{i}(s)$:
\begin{equation*}
	\tilde{\omega}_{i}(\tau) = \frac{ v_{s,i}(\tilde{\rho}_i, \tilde{\varphi}_i)}{v_{s,max,i}} \cdot \bar{\omega}_{i}(s(\tau))
\end{equation*}
which ensures the trial trajectory passes over predicted path and preserves the same curvature, but not producing the same length, regardless of $\tilde{\rho}$ and $\tilde{\varphi}$ being employed.

Vertical component of a trial trajectory is also determined by the trial regimes $\tilde{\rho}$ and $\tilde{\varphi}$:
\begin{equation*}
	\frac{\diff\tilde{z}_i}{\diff\tau} = v_{z,i}(\tilde{\rho}_i, \tilde{\varphi}_i).
\end{equation*}

Trial velocity intensity and incline angle function, $\tilde{\rho}(\tau)$ and $\tilde{\varphi}(\tau)$, respectively, are subjected to the optimization in the model predictive control framework, unlike trial yaw angular velocity $\tilde{\omega}$ that is obtained from the gradient of the current potential field and corrected according to $\tilde{\rho}(\tau)$ and $\tilde{\varphi}(\tau)$.

\subsection{Optimization problem formulation}
We seek for an optimal trial flight regime in time window $[t, t+ \tau_{max}]$ in order to determine velocity and altitude control regulated by $\rho$ and $\varphi$ at time $t$. 
In order to assemble the optimization problem, $\tilde{\rho}$ and $\tilde{\varphi}$ need to be parameterized.
These functions are obtained through quadratic polynomial interpolation defined with three specific points: $\tau_0 = 0$ (known current UAV state at time $t$), $\tau_1 = \frac{\tau_{max}}{2}$, and $\tau_2 = \tau_{max}$. 
While trial control functions are know at (current) time $t$:
\begin{equation*}
	\begin{split}
		\tilde{\rho}(\tau_0) &= \rho(t)\\
		\tilde{\varphi}(\tau_0) &= \varphi(t).
	\end{split}
\end{equation*}
we define the optimization vector $\mathbf{W}\in\mathbb{R}^4$ as values of $\tilde{\rho}$ and $\tilde{\varphi}$ at $\tau_1$ and $\tau_2$:
\begin{equation*}
	\mathbf{W}_i \equiv \left[\tilde{\rho}_i(\tau_1),~ \tilde{\rho}_i(\tau_2),~ \tilde{\varphi}_i(\tau_1),~ \tilde{\varphi}_i(\tau_2) \right]^T.
\end{equation*}

Since the trial control functions are functions of relative time $\tau$ but in terms of the optimization, they are also defined by (functions of) the optimization vector $\mathbf{W}_i$, for the sake of simplicity and readability, we introduce the notation:
\begin{equation*}
	(\cdot)\big\lfloor_{\mathbf{W}_i}(\tau)
\end{equation*}
which means that any trial function $(\cdot)$ is function of relative time $\tau$ and optimization vector $\mathbf{W}_i$.

We set two objectives for the flight regime: maximizing UAV velocity and providing the UAV altitude closest possible to the goal search altitude $h_{goal}$. Maximizing UAV velocity eventually leads to shorter inspection time, and this is accomplished with introduction of the minimization objective:
\begin{equation}
	o_{v,i}(\mathbf{W}_i) = 1 - \frac{1 }{\tau_{max}} \int_{0}^{\tau_{max}} \tilde{\rho_i}\big\lfloor_{\mathbf{W}_i}(\tau) \diff\tau.
	\label{eq:velocity_objective}
\end{equation}

To define second objective we introduce relative terrain height function 
\begin{equation*}
	\tilde{z}_{T,i}\big\lfloor_{\mathbf{W}_i}(\tau) = z_T\left(\tilde{x}_i\big\lfloor_{\mathbf{W}_i}(\tau),~ \tilde{y}_i\big\lfloor_{\mathbf{W}_i}(\tau)\right).
\end{equation*}

The proximity of the trial altitude function $\tilde{z}-\tilde{z}_T$ to the goal altitude $h_{goal}$ is determined by the altitude objective 
\begin{equation}
	o_{h,i}(\mathbf{W}_i) = \frac{ \int_{0}^{\tau_{max}}  \left| \tilde{z}_i\big\lfloor_{\mathbf{W}_i}(\tau) -  \tilde{z}_{T,i}\big\lfloor_{\mathbf{W}_i}(\tau) - h_{goal,i} \right|  \diff\tau
	}{h_{goal,i}\cdot\tau_{max}}.
	\label{eq:goal_altitude_objective}
\end{equation}

Both velocity and altitude objectives are normed by their definitions \eqref{eq:velocity_objective} and \eqref{eq:goal_altitude_objective}, respectively. Hence the two objectives are combined by addition, without using weight factors since they have the same significance:
\begin{equation*}
	o_i(\mathbf{W}_i) = o_{v,i}(\mathbf{W}_i) + o_{h,i}(\mathbf{W}_i)
\end{equation*}

We define the constraints of a trial trajectory in the non-equality form $c(\mathbf{W}) \leq 0$.
In order to guarantee that the UAV cannot go below specified minimum altitude $h_{min}$ we impose a minimum altitude optimization constraint 
\begin{equation}
	\tilde{z}_i\big\lfloor_{\mathbf{W}_i}(\tau) -  \tilde{z}_{T,i}\big\lfloor_{\mathbf{W}_i}(\tau) \geq h_{min,i}.
	\label{eq:min_altitude_constraint}
\end{equation}
It needs to be viable for every $\tau$ which is validated using
\begin{multline*}
	c_{h,i}(\mathbf{W}_i) = \\
	\frac{ \int_{0}^{\tau_{max}} \max\left\{ h_{min,i} - \tilde{z}_i\big\lfloor_{\mathbf{W}_i}(\tau) +  \tilde{z}_{T,i}\big\lfloor_{\mathbf{W}_i}(\tau), ~0\right\}
		\diff\tau
	}{h_{min,i}\cdot\tau_{max}} .
\end{multline*}

To ensure compliance with the velocity and acceleration specification, the following constraints need to be satisfied:
\begin{equation}
	v_{s,min,i} \leq \tilde{v}_{s, i}\big\lfloor_{\mathbf{W}_i}(\tau) \leq v_{s,max,i}
	\label{eq:vs_constraints}
\end{equation}
\begin{equation}
	v_{z,min,i} \leq \tilde{v}_{z, i}\big\lfloor_{\mathbf{W}_i}(\tau) \leq v_{z,max,i}
	\label{eq:vz_constraints}
\end{equation}
where $\tilde{v}_{s, i}\big\lfloor_{\mathbf{W}_i}(\tau)$ and $\tilde{v}_{z, i}\big\lfloor_{\mathbf{W}_i}(\tau)$ are horizontal and vertical trial velocity functions obtained with \eqref{eq:v_s} and \eqref{eq:v_z}, respectively.
Evaluation of horizontal velocity constraints is performed with:
\begin{equation*}
	c_{vs min,i}(\mathbf{W}_i) =  \int_{0}^{\tau_{max}}
	\frac{ \max\left\{ v_{s,min,i} - \tilde{v}_{s, i}\big\lfloor_{\mathbf{W}_i}(\tau), ~0\right\}
		\diff\tau
	}{v_{s,min,i}\cdot\tau_{max}}
\end{equation*}
\begin{equation*}
	c_{vs max,i}(\mathbf{W}_i) =  \int_{0}^{\tau_{max}}
	\frac{ \max\left\{\tilde{v}_{s, i}\big\lfloor_{\mathbf{W}_i}(\tau) - v_{s,max,i}, ~0\right\}
		\diff\tau
	}{v_{s,max,i}\cdot\tau_{max}}
\end{equation*}
and analogously for vertical velocity constraints:
\begin{equation*}
	c_{vz min,i}(\mathbf{W}_i) =  \int_{0}^{\tau_{max}}
	\frac{ \max\left\{ v_{z,min,i} - \tilde{v}_{z, i}\big\lfloor_{\mathbf{W}_i}(\tau), ~0\right\}
		\diff\tau
	}{v_{z,min,i}\cdot\tau_{max}}
\end{equation*}
\begin{equation*}
	c_{vz max,i}(\mathbf{W}_i) =  \int_{0}^{\tau_{max}}
	\frac{ \max\left\{\tilde{v}_{z, i}\big\lfloor_{\mathbf{W}_i}(\tau) - v_{z,max,i}, ~0\right\}
		\diff\tau
	}{v_{z,max,i}\cdot\tau_{max}}.
\end{equation*}

The acceleration of UAV is also considered in trial trajectories. Horizontal and vertical accelerations depend on properties and technical abilities of the UAV, such as vehicle mass and inertia, and achievable upthrust and pitch in low-level UAV control. The most simple acceleration and deceleration constraints are imposed to horizontal and vertical UAV motion:
\begin{equation}
	a_{s,min,i} \leq \tilde{a}_{s, i}\big\lfloor_{\mathbf{W}_i}(\tau) \leq a_{s,max,i}
	\label{eq:as_constraints}
\end{equation}
\begin{equation}
	a_{z,min,i} \leq \tilde{a}_{z, i}\big\lfloor_{\mathbf{W}_i}(\tau) \leq a_{z,max,i}.
	\label{eq:az_constraints}
\end{equation}
Acceleration trial functions $\tilde{a}_{s, i}\big\lfloor_{\mathbf{W}_i}(\tau)$ and $\tilde{a}_{z, i}\big\lfloor_{\mathbf{W}_i}(\tau)$ are calculated via numerical differentiation of corresponding horizontal and vertical velocities $ \tilde{v}_{s, i}\big\lfloor_{\mathbf{W}_i}(\tau)$ and $\tilde{v}_{z, i}\big\lfloor_{\mathbf{W}_i}(\tau)$, respectively.

Horizontal acceleration and deceleration constraints are evaluated with:
\begin{equation*}
	c_{as min,i}(\mathbf{W}_i) =  \int_{0}^{\tau_{max}}
	\frac{ \max\left\{ a_{s,min,i} - \tilde{a}_{s, i}\big\lfloor_{\mathbf{W}_i}(\tau), ~0\right\}
		\diff\tau
	}{a_{s,min,i}\cdot\tau_{max}}
\end{equation*}
\begin{equation*}
	c_{as max,i}(\mathbf{W}_i) =  \int_{0}^{\tau_{max}}
	\frac{ \max\left\{\tilde{a}_{s, i}\big\lfloor_{\mathbf{W}_i}(\tau) - a_{s,max,i}, ~0\right\}
		\diff\tau
	}{a_{s,max,i}\cdot\tau_{max}}
\end{equation*}
and analogously constraints for vertical acceleration and deceleration:
\begin{equation*}
	c_{az min,i}(\mathbf{W}_i) =  \int_{0}^{\tau_{max}}
	\frac{ \max\left\{ a_{z,min,i} - \tilde{a}_{z, i}\big\lfloor_{\mathbf{W}_i}(\tau), ~0\right\}
		\diff\tau
	}{a_{z,min,i}\cdot\tau_{max}}
\end{equation*}
\begin{equation*}
	c_{az max,i}(\mathbf{W}_i) =  \int_{0}^{\tau_{max}}
	\frac{ \max\left\{\tilde{a}_{z, i}\big\lfloor_{\mathbf{W}_i}(\tau) - a_{z,max,i}, ~0\right\}
		\diff\tau
	}{a_{z,max,i}\cdot\tau_{max}}
\end{equation*}
respectively.

\subsection{Solving MPC optimization}
\begin{figure}[!ht]
	\centering
	\includegraphics[width=0.95\linewidth]{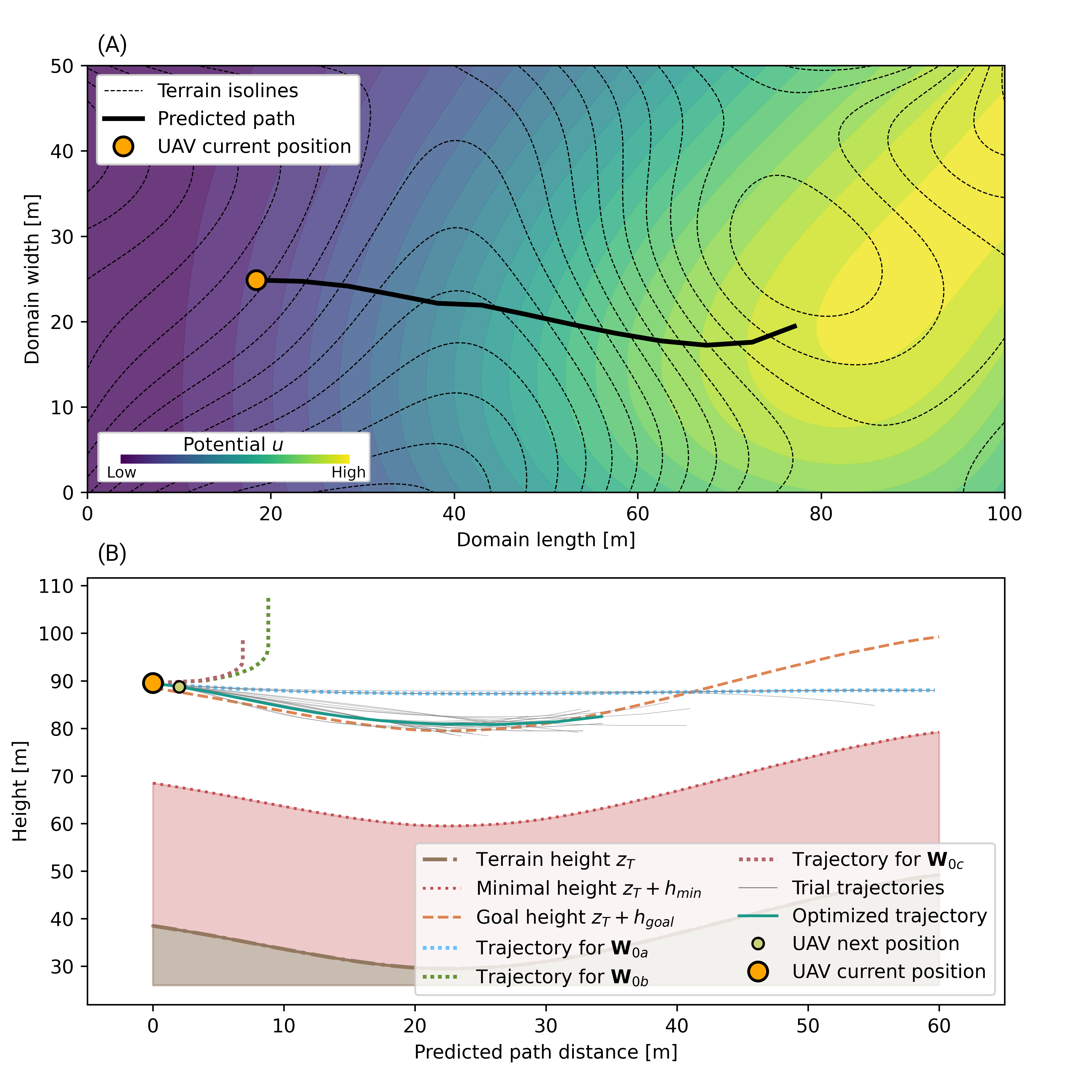}
	\caption{Visual representation of path prediction and the optimization process. The predicted path is formed starting from the current position of the UAV, simulating maximal achievable horizontal movement for the number of prediction steps $n_{pts}$, along the gradient of the potential field $u$ as shown in (A). Terrain height $z_T$ is then probed along the length of the predicted path and used to construct minimal ($z_T + h_{min}$) and goal ($z_T + h_{goal}$) height curves. The optimization process, depicted in (B), initiates at the first feasible regime among $\mathbf{W}_{0a}, \mathbf{W}_{0b}, \mathbf{W}_{0c}$, following the order as listed. It balances between velocity maximization and goal altitude adherence objectives to generate the optimal flight regime for $n_{pts}$ time steps while taking velocity, acceleration and minimal height constraints into consideration. The UAV then utilizes the optimized flight regime for $\Delta t$ seconds, consequently, it moves to the new position (marked with a lime green colored dot) where the complete process repeats.}
	\label{fig:path_predictor}
\end{figure}

Each optimization is started with a specific initial optimization vector $\mathbf{W}_0$ which is chosen between:   
\begin{equation*}
	\mathbf{W}_{0a} = (1, 1, 0, 0),
\end{equation*}
\begin{equation*}
	\mathbf{W}_{0b}  = (1, 1, 1, 1),
\end{equation*}
\begin{equation*}
	\mathbf{W}_{0c}  = (0.5, 0.5, 1, 1).
\end{equation*}
We check possible initial vectors in order from $a$ to $c$ and the first one that returns feasible solution is selected as $\mathbf{W}_0$. $\mathbf{W}_{0a}$ encourages the utilization of maximal forward velocity in the horizontal (search) direction while $\mathbf{W}_{0b}$ and $\mathbf{W}_{0c}$ advocate increase in altitude with different velocity intensity. Figure \ref{fig:path_predictor} displays an illustration of the optimization process and possible initial optimization vectors. If there is no feasible solution, UAV skips the optimization process and starts collision avoidance procedure (described below). 

Optimization is performed using a variant of Generalized Pattern Search (GPS) method called Multi-Scale Grid Search (MSGS) implemented in \textit{Indago} Python module \cite{indago}. Number of optimization iterations is set to 30 while the stopping criteria are regulated by setting the maximum number of stalled iterations to 10 and target fitness to $10^{-3}$. Using the optimized values $\mathbf{W}_{opt}$ we obtain the candidate control values for velocity intensity:
\begin{equation*}
	\rho_{opt, i} = \tilde{\rho}_i\big\lfloor_{\mathbf{W}_{opt,i}}(\Delta t), 
\end{equation*}
incline:
\begin{equation*}
	\varphi_{opt, i} = \tilde{\varphi}_i\big\lfloor_{\mathbf{W}_{opt,i}}(\Delta t)
\end{equation*}
and omega:
\begin{equation*}
	\omega_{opt, i} = \tilde{\omega}_i\big\lfloor_{\mathbf{W}_{opt,i}}(\Delta t).
\end{equation*}

\subsection{Validating escape maneuver and establishing control parameters}

Although optimal flight control parameters $\rho_{opt,i}$, $\varphi_{opt,i}$ and $\omega_{opt,i}$ are determined in MPC optimization, they do not guarantee a possibility of collision-free trajectory (realized with escape maneuver) for the next time step. Due to high computational demands, considering this constraint in MPC optimization is not feasible for achieving real-time UAV control. Hence, the obtained MPC flight control parameters need to be verified and corrected if needed before they are enforced to UAVs.

The escape maneuver is a specific UAV motion that aims to provide following flight features:
\begin{itemize}
	\item Escape maneuver can be started from any starting time given any locations, state and control parameters of the UAV. Here we consider UAVs at time $t+\Delta t$ with optimal control parameters and, if needed, at current time $t$ with current control parameters.
	\item The projection of the escape trajectory to the horizontal plane is circular arc (extendable to full circle) with radius equal to $\pm R_{min,i}$, while adjusting $\omega_{esc, i}$. This ensures the resulting arc overlaps with one of the escape circles defined in section~\ref{subsec:hedac_collison_avoidnace}.
	\item The UAV decelerate in the most agile feasible regime i.e. $\rho_{em, i} \rightarrow \rho_{min, i}$.
	\item The UAV ascends in the most agile feasible regime i.e. $\varphi_{em, i} \rightarrow \varphi_{max, i}$.
	\item Escape route feasibility is determined by before-mentioned constraints: minimal altitude \eqref{eq:min_altitude_constraint}, horizontal and vertical velocity \eqref{eq:vs_constraints} and \eqref{eq:vz_constraints}, horizontal and vertical acceleration \eqref{eq:as_constraints} and \eqref{eq:az_constraints}.
	\item The considered duration of the escape motion is until the horizontal velocity drops to zero or until full escape circle is passed.
\end{itemize}

If a feasible escape maneuver is achievable for optimal control parameters starting at time $t+\Delta t$, the  optimal control parameters can be applied to UAV:
\begin{equation*}
	\rho_i (t + \Delta t) = \rho_{opt,i}
\end{equation*}
\begin{equation*}
	\varphi_i (t + \Delta t) = \varphi_{opt,i}
\end{equation*}
\begin{equation*}
	\omega_i (t + \Delta t) = \omega_{opt,i}.
\end{equation*}

If there is no feasible escape maneuver at time $t + \Delta t$, indicating currently obtained control parameters are not valid, the UAV immediately starts the escape maneuver using parameters obtained at $t+\Delta t$ (for verified escape maneuver simulated from time $t$):
\begin{equation*}
	\rho_i (t + \Delta t) = \rho_{em,i}(t+\Delta t)
\end{equation*}
\begin{equation*}
	\varphi_i (t + \Delta t) = \varphi_{em,i}(t+\Delta t)
\end{equation*}
\begin{equation*}
	\omega_i (t + \Delta t) = \omega_{em,i}(t+\Delta t).
\end{equation*}

\section{Tested search scenarios}

In order to test the method, we designed three test cases with different terrain complexity and size. Each case uses a specific configuration of different UAVs, all of which are described in Table \ref{tab:uav_charactristics}. 

\begin{table}[h!]
	\scriptsize
	\centering
	\caption{Motion, vision/sensing and control UAV parameters used in simulated search scenarios}
	\begin{tabularx}{\linewidth}{L{2.95}R{0.6}R{0.6}R{0.6}L{0.25}} 
		UAV parameters	& UAV A 		& UAV B		 	& UAV C			& Units				\\
		\hline 	
		Type  												& Multi-rotor	& Multi-rotor 	& Fixed-Wing	& -					\\
		Minimum turning radius $R_{min}$							& 25			& 25			& 100			& m					\\
		Minimum clearance distance $\delta$					& 7 			& 7				& 60			& m					\\
		Minimum search altitude	$a_{min}$					& 30			& 30			& 100			& m					\\
		Goal search altitude $a_{goal}$ 					& 50			& 100			& 150			& m					\\
		Maximum horizontal velocity $v_{s_{max}}$ 			& 10			& 10			& 15			& m/s				\\
		Minimum horizontal velocity $v_{s_{min}}$ 			& 0				& 0				& 5				& m/s				\\
		Maximum ascending velocity $v_{z_{max}}$ 			& 5				& 5				& 1.2			& m/s				\\
		Maximum descending velocity $v_{z_{min}}$ 			& -3			& -3			& -1.2			& m/s				\\
		
		Maximum horizontal acceleration $v_{s_{max}}$ 		& 2				& 2				& 2				& m/s				\\
		Minimum horizontal acceleration $v_{s_{min}}$ 		& -3.6			& -3.6 			& -2			& m/s				\\
		Maximum vertical acceleration $v_{z_{max}}$ 		& 2.8			& 2.8				& 1				& m/s				\\
		Minimum vertical acceleration $v_{z_{min}}$ 		& -2			& -2			& -1 			& m/s				\\
		
		Minimum incline $\varphi_{min}$						& -90			& -90			& 13.5			& $^\circ$			\\
		Maximum incline $\varphi_{max}$						& 90			& 90			& 13.5			& $^\circ$			\\
		
		Camera FOV ($\gamma_1$, $\gamma_2$) 				& (62.8, 37.9) 	& (44.2, 21.3)	& (90, 54.3)	& $^\circ$			\\
		
		Sensing function $\Gamma$							& $\Gamma_A$ 	& $\Gamma_B$ 	& $\Gamma_C$	& -					\\
		
		Predicted path length $n_{pts}$			 			& 20 			& 20			& 30			& time steps		\\
	\end{tabularx}
	\renewcommand{\arraystretch}{1.6}
	\begin{tabularx}{\linewidth}{Xl} 
		Sensing functions 							& 				\\
		\hline		
		$\Gamma_A$						& $1.2 \cdot \text{exp} ((-50 - d)/25)^{0.55}$ \\
		$\Gamma_B$						& $1.4 \cdot \text{exp} ((-45 - d)/35)^{0.6}$  \\
		$\Gamma_C$						& $1.6 \cdot \text{exp} ((-45 - d)/35)^{0.6}$  \\
	\end{tabularx}
	\label{tab:uav_charactristics}
\end{table}

Multi-rotor UAVs (A, B) are modeled based on technical specifications and tests performed with \textit{DJI Matrice 210 V2} aircraft. We carried out an experiment to confirm or conclude maximal horizontal, ascending and descending velocity and minimal and maximal value of horizontal and vertical acceleration which we used. It's important to note that despite both UAVs being modeled similarly, they were regulated to conduct searches at varying altitudes according to their technical characteristics. The deliberate difference in altitude between A, employing a goal altitude of 50 meters above ground level, and B, utilizing a goal altitude of 100 meters while incorporating and simulating camera zoom, leads to a unique FOV and sensing function combination for each UAV type. Characteristics of the fixed-wing UAV (C) are roughly estimated to portray a realistic aircraft which is significantly more constrained compared to multi-rotor UAVs. Its design targets the use in large domains with lower incline values and smoother terrain.

\begin{figure}[!htb]
	\centering
	\includegraphics[width=\linewidth]{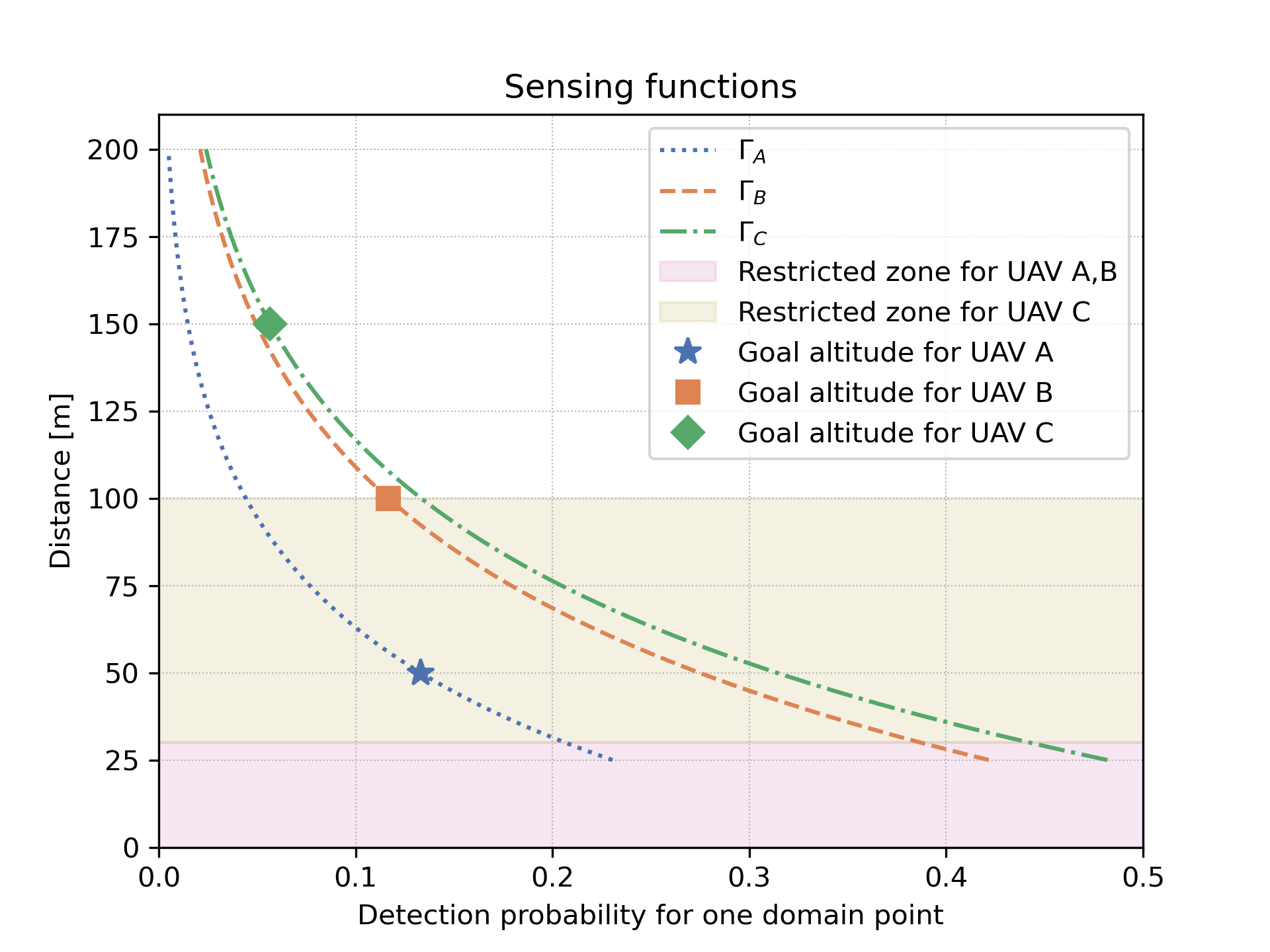}
	\caption{Relationship between sensing functions and the parameters of UAVs that employ them. When considering the sensing functions, the y axis represents the distance between the sensor and the observed point, but when examining the restricted flight zones and goal altitudes it represents the distance between the sensor (or the UAV) to the point directly below the UAV (that distance is equal to its altitude).}  
	\label{fig:sensing_functinos}
\end{figure}

For each camera sensor that was used, we fabricated an accompanying sensing function which correspond to the accuracy of realistic detection model. Cameras featuring a broader FOV encompass a larger area at the same altitude, resulting in fewer pixels capturing the search target. Consequently, this reduction leads to a diminished detection probability rate, so the sensing functions are adjusted accordingly. Figure \ref{fig:sensing_functinos} displays all the sensing functions used and associates them with the altitude goals and constraints for specific UAVs that utilize them. 

For conducting search simulations, we generated two-dimensional meshes with triangular elements using \textit{Gmsh} \cite{geuzaine2009gmsh}. Each node within the mesh includes terrain elevation data. Domain and mesh details can be found in Table \ref{tab:test_cases} along with method parameters and number of specific UAVs used. 

\begin{table}[h!]
	\scriptsize
	\centering
	\caption{Test cases parameters and data}
	\begin{tabularx}{\linewidth}{L{2.95}R{0.6}R{0.6}R{0.6}L{0.25}} 
		Test case parameters & Plastic world		& Mt. \newline Vesuvius 		& Star dunes 		& Units				\\
		\hline 			
		Domain size 										& 0.72				& 7.44 				& 7.5				&  $\text{km}^2$	\\
		Number of mesh nodes								& 8380				& 21825				& 21946				& -					\\
		Number of mesh elements 							& 16300				& 43098 			& 43340				& -					\\
		Elevation difference		& 421				& 608				& 221.4				& m					\\
		Alpha $\alpha$ 										& 1000				& 500  				& 500				& -					\\
		Beta $\beta$ 										& 0.1				& 0.4 				& 0.1				& -					\\
		Time step $\Delta t$								& 1					& 1 				& 2					& s					\\
		Search duration										& 15				& 60				& 60				& min				\\
		Number of UAVs A									& 3					& 3 				& 0					& -					\\
		Number of UAVs B									& 0					& 2					& 0					& - 				\\
		Number of UAVs C									& 0					& 0					& 2 				& -					\\
	\end{tabularx}
	\label{tab:test_cases}
\end{table}

Simulations were performed on a machine with 3.7 GHz base CPU clock. Array manipulation along with other numerical tools form \textit{NumPy} \cite{harris2020array} and interpolation tools from \textit{SciPy} \cite{2020SciPy-NMeth} were used. Test cases visualization and analysis is performed using \textit{Matplotlib} \cite{Hunter:2007} and \textit{PyVista} \cite{sullivan2019pyvista}.

\begin{figure*}[h!]
	\centering
	\includegraphics[width=1\linewidth]{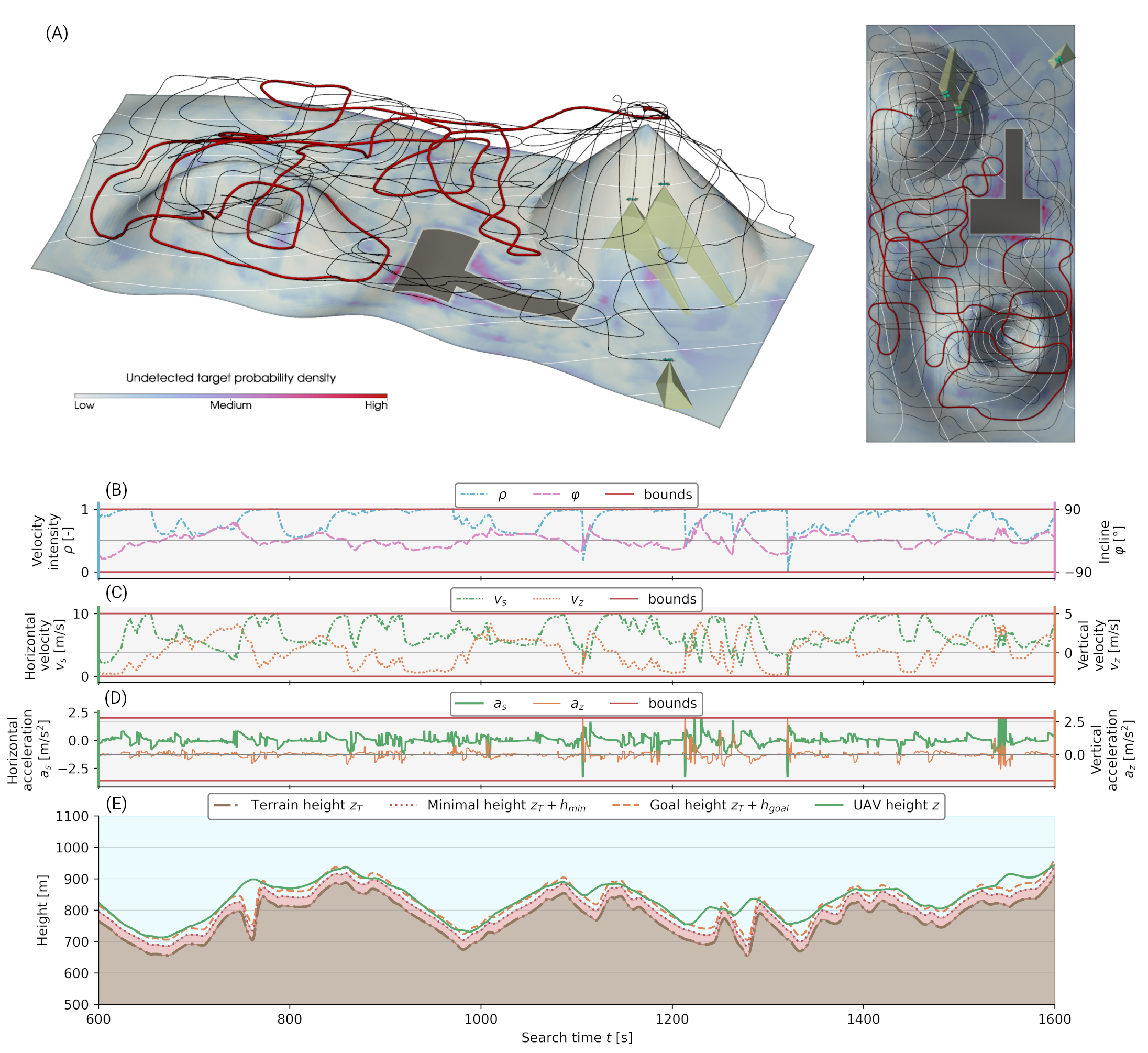}
	\caption{Plastic world case survey simulation after 30 minutes of search time including UAVs' trajectories (A) with an in depth  parameter analysis for the trajectory marked in red. High aircraft maneuverability is deducted from frequent variations of control parameters (B) which lead to velocity (C) and acceleration (D) jumps. Height analysis of the red trajectory which is performed by an UAV with a goal altitude of 50 m and a minimum altitude of 30 m is shown in (E). The UAV flies relatively closely to the goal altitude except upon detecting elevation jumps (shown at about 700, 1100 and 1200 s) where it tends to promptly ascend and when flying over terrain depressions it sometimes chooses to favor velocity over altitude (shown at about  770 s and 1280 s) keeping its momentum while the resulting trajectory is smooth. The inclination for terrain adherence and velocity maximization is evident on elevation peaks, where altitude drops below the goal value (sometimes even up to minimal altitude) and then after crossing the peak UAV ascends back to goal value which results in smoother trajectories, allowing for faster flight.} 
	\label{fig:pw_3d_ta}
\end{figure*}

The first test is performed on a synthetically generated domain which we called "Plastic world". Enveloped terrain is purposely constructed to be simple and exaggerated, yet providing specific features of natural terrain, in order to demonstrate the robustness of the algorithm. Search is conducted collaboratively using 3 multi-rotor UAVs. Search domain and accompanying UAV trajectories after survey completion, in addition with the analysis of control parameters and the associated trajectory performed by one UAV during a 1000 s time window, are presented in Figure \ref{fig:pw_3d_ta}. After evaluating the trajectory, it is evident that all control and flight parameter constraints have been fulfilled. Depending on the horizontal moving velocity and undetected target probability, the UAV decides whether to slow down and inspect the valley closer or fly over at slightly larger altitude without sacrificing its horizontal moving velocity (and slowing down the search process). Upon detecting a sudden increase in terrain elevation, it tends to ascend rapidly, reducing detection probability but increasing safety distance which is especially beneficial if dealing with substandard terrain elevation models (coarsely sampled). The rectangular no-fly zone is avoided by all UAVs respecting assigned clearance distance wile the area closely surrounding it has been successfully explored.

\begin{figure*}[h!]
	\centering
	\includegraphics[width=0.95\linewidth]{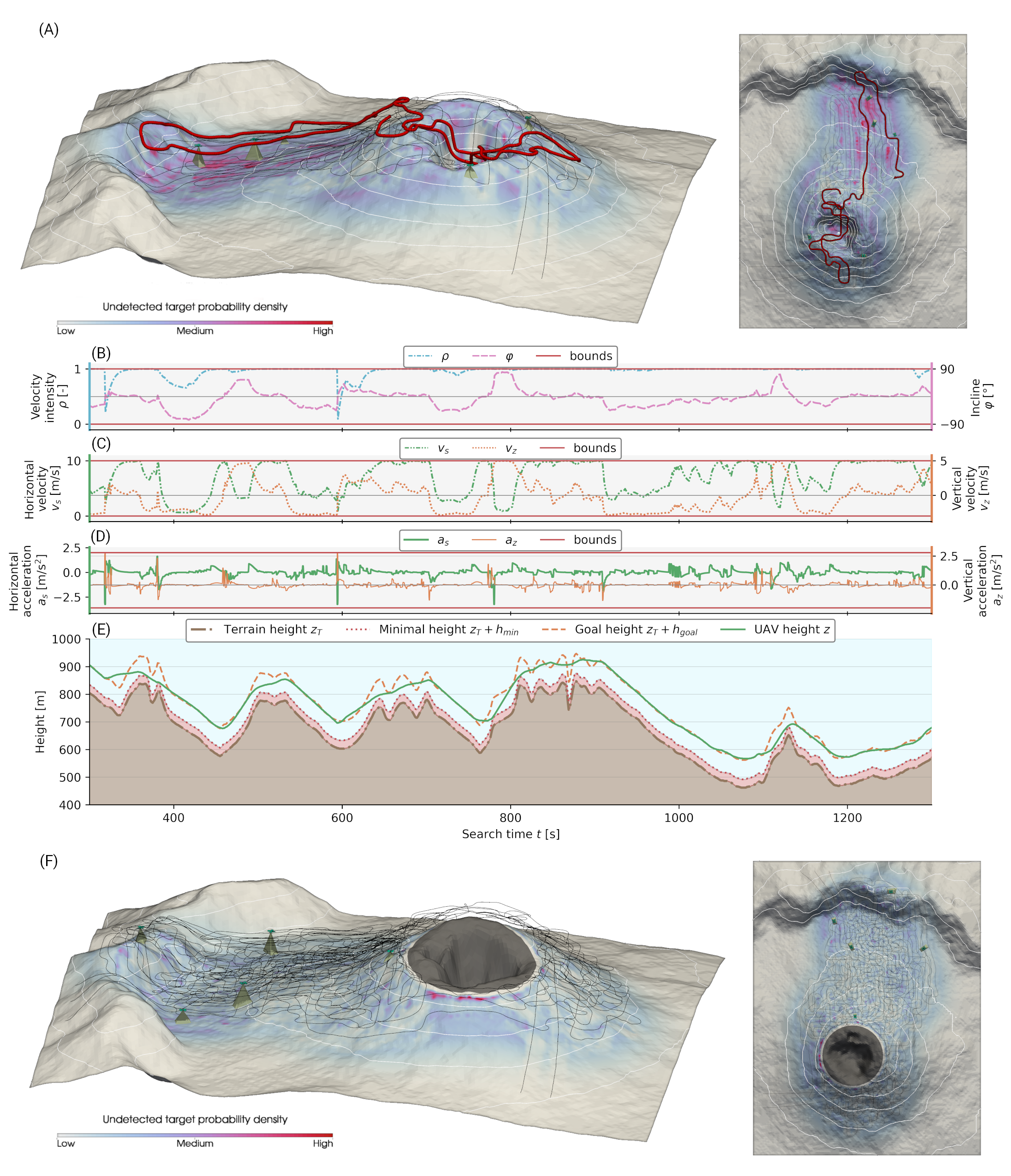}
	\caption{Mount Vesuvius case survey simulation with 5 cooperating UAVs after 1400 seconds. The 3D scene (A) contains the terrain colored in reference to the undetected target probability with gray representing the lowest and red representing the highest value. The black lines represent UAVs' trajectories and the red highlighted part, which is executed by an UAV with a goal altitude of 100 m and a minimum altitude of 30 m, is analyzed in graphs (B-E). It is clear that the control parameters, velocities and accelerations (B-D) stay within the prescribed limits and correspond to each other. Graph (E) showcases an excellent example of trajectory smoothing and altitude variation dampening effect at elevation peaks (at about 370, 450, 700 and 850 s). Rather than strictly adhering to the goal altitude when encountering consecutive elevation peaks, the UAV efficiently navigates through these peaks, maintaining high velocity and achieving a remarkably smooth trajectory. When flying over a single altitude peak, as shown around 1150 seconds, the UAV generally exploits the full permitted altitude range, descending to the minimum allowed altitude at the peak's center and then ascending back to goal altitude. This results in a notably smoother trajectory, enhancing flight efficiency. Survey with incorporated no-fly zone after 60 minutes of search is shown in (F). When compared to the unrestricted case (A), where UAVs fly in and out the volcano crater, we can notice the effect of the restricted zone on the final trajectories and UAVs' compliance to the restriction.} 
	\label{fig:mv_3d_ta}
\end{figure*}

\begin{figure}[!ht]
	\centering
	\includegraphics[width=\linewidth]{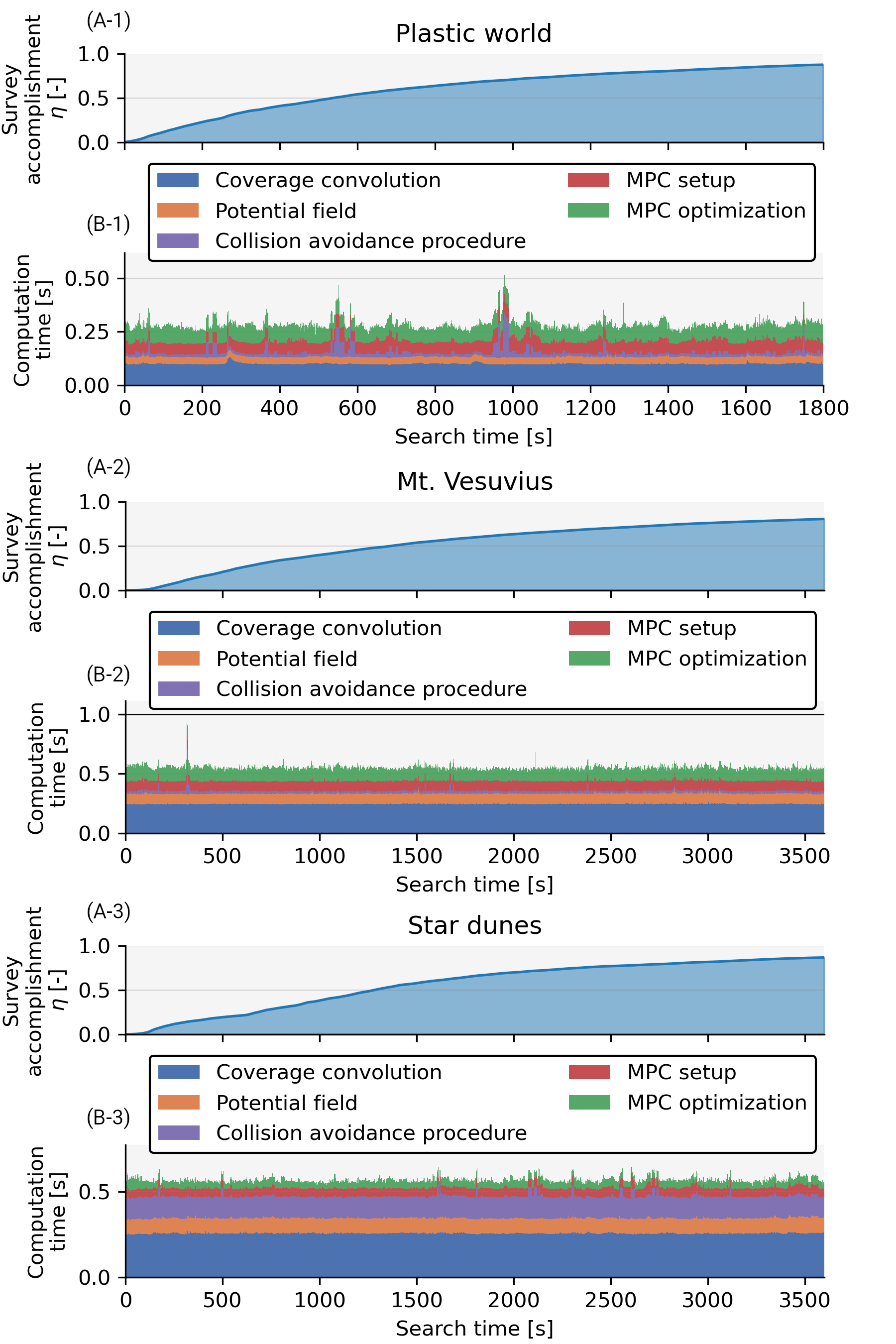}
	\caption{Survey accomplishment and computation time for executed tests. After concluding the search, all cases attained a survey accomplishment metric exceeding 80\% (A) based on selected configurations, including search time, HEDAC parameters, time step, and defined UAV constraints. The computation time for each individual time step shorter than the control time step $\Delta t$ ($\Delta t =$ 1 s for "Plastic world" and "Mount Vesuvius" cases using multi-rotors and $\Delta t =$ 2 s for "Star dunes" using fixed-wing UAVs) implies the suitability for real-time applications of proposed multi-UAV search control algorithm (B). The computation time peaks are the consequence of prolonged collision avoidance procedure (for example, at around 1000 s on B-1) which occurs when multiple UAVs are nearby. The number of UAVs, domain size and initial target probability distribution effect the frequency and intensity of collision avoidance procedure time jumps.}
	\label{fig:cptime}
\end{figure}

The second test case is a survey simulation of Mount Vesuvius which is situated near Naples, Italy. We chose to survey this area only with multi-rotor UAVs because of their high maneuverability, hence they could easily search volcanic cone and crater. In this case we demonstrated non-homogeneous swarm support by using 3 UAVs flying at an altitude of 50 m with one type of camera sensor and 2 UAVs flying at an altitude of 100 m were we simulated the use of camera zoom (the sensor has narrower FOV and higher-detection sensing function was used). The search mission was simulated for 60 minutes. The inspection was completed with sufficient survey accomplishment metric, which is evident on Figure \ref{fig:cptime}, without any collisions or breaching velocity, acceleration or altitude constraints. The proposed motion control successfully balances between maximizing velocity and ideal altitude objectives while complying all constraints. Due to more natural and smooth terrain, compared to the Plastic world case, the UAV is able to keep its maximum velocity almost the entire flight.

To demonstrate a realistic search scenario involving a restricted area, we conducted another survey simulation of Mount Vesuvius with the identical case configuration, the only difference between the cases is that we declared the volcano crater a no-fly zone. Visualizations in Figure \ref{fig:mv_3d_ta}(A-E) and \ref{fig:mv_3d_ta}(F) exhibits simulation results with and without the no-fly zone, respectively. It showcases the terrain, undetected target probability distribution and UAVs' trajectories, along with a detailed analysis of a selected part of the UAV flight. Restricting the area of the search does not have a significant qualitative impact on exploration, although, due to the excluded surface area, a more detailed search of the remaining zones is achieved in the same amount of time.

\begin{figure*}[h!]
	\centering
	\includegraphics[width=1\linewidth]{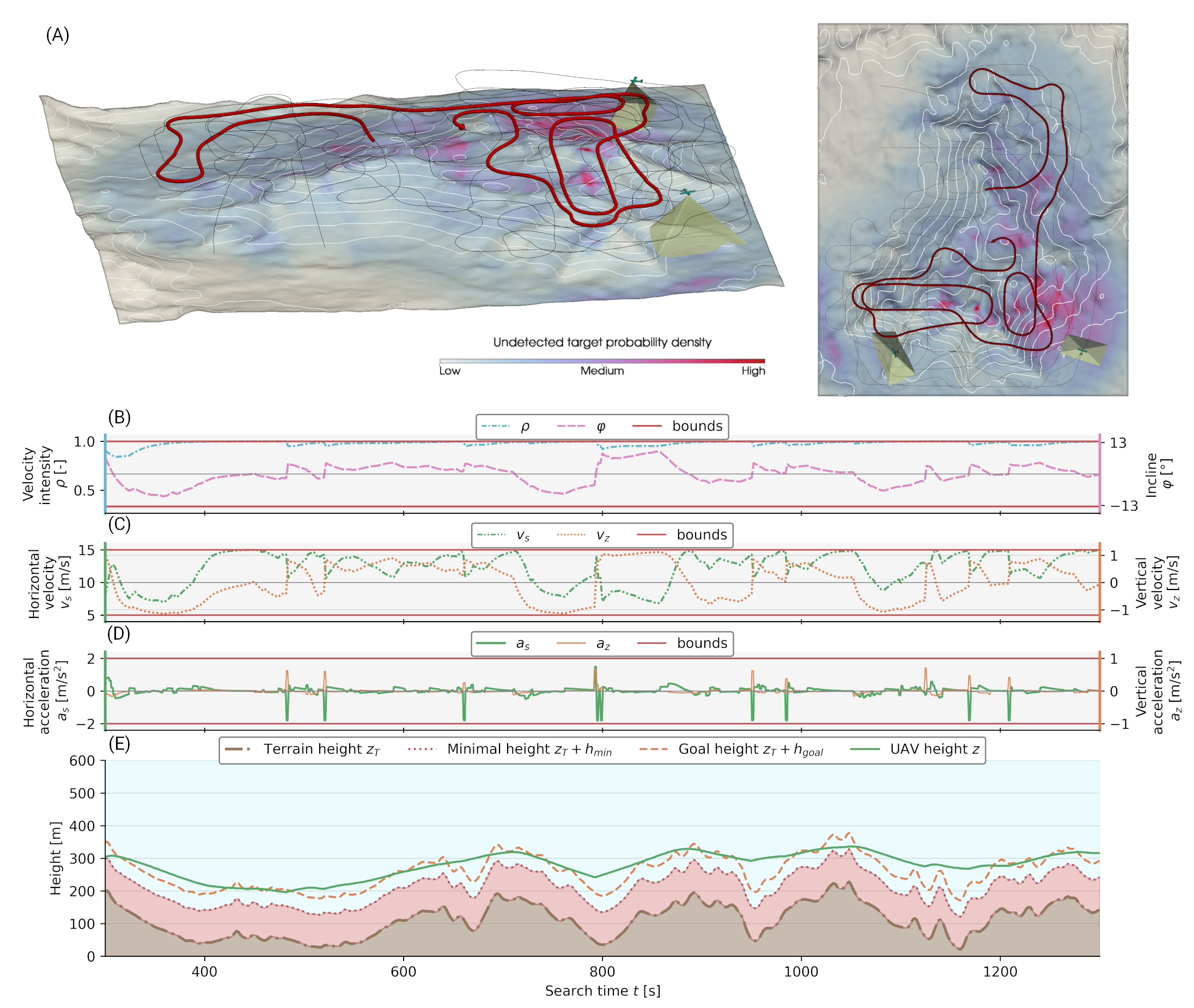}
	\caption{Star dunes case inspection after 60 minutes. 3D scene (A) displays the terrain with colors matching undetected target probability, with gray being the lowest and red the highest value. Black lines depict aircraft trajectories and the highlighted red trajectory section is further analyzed in graphs (B-E). As evident from control parameters graph (B), the aircraft tends to keep its moving intensity at the maximum, lowering it slightly when changing pitch, in order to satisfy acceleration constraints. Velocity and acceleration values (C-D) are corresponding to the control parameters and their values are inside prescribed limits. During its flight, the aircraft attempts to to closely track the target altitude. However, due to its operational constraints, the resulting trajectory exhibits a strong dampening effect on terrain elevation irregularities, making it much more fluid, allowing the aircraft to effortlessly navigate over terrain obstacles. } 
	\label{fig:sd_3d_ta}
\end{figure*}

For the third test, we performed a search simulation of a sandy desert in Algeria. We took a 7.5 km$^2$ area that contains star dunes, which are dunes that form in sandy deserts where the wind direction frequently changes. They can grow to considerable heights and are generally taller than other types of sand dunes, which is why they were chosen for altitude control demonstration. The search was conducted for 60 minutes, but since we employed fixed-wing UAVs, which have slower responses and do not need as frequent commands, the control time step was set to 2 seconds. Due to their higher velocity, greater altitude and  bigger overall area coverage we achieved adequate survey accomplishment (Figure \ref{fig:cptime}) by using only 2 UAVs, compared to the Mount Vesuvius case where we used 5 multi-rotor UAVs to inspect an area of almost identical size. Figure \ref{fig:sd_3d_ta} illustrates the search domain, the UAVs, their post-survey trajectories, and an analysis of the highlighted trajectory displaying 1000 s of flight time. The analyzed trajectory covers nearly the entire domain. During its flight, the UAV prioritizes maximizing velocity while making altitude adjustments in order to stay close to the target altitude and avoid the minimal altitude zone.

Survey accomplishment and computation time analysis were performed for all 3 cases and are shown in Figure \ref{fig:cptime}. With the selected UAV configurations and HEDAC parameters, shown in Table \ref{tab:test_cases}, all the cases achieved more than 80\% survey accomplishment in the specified search time. Computation time for each individual time step is lower than the time step $\Delta t$ used in the calculation which is a good indication for successful real-world application.

\section{Possible drawbacks}

\begin{figure}[!ht]
	\centering
	\includegraphics[width=0.9\linewidth]{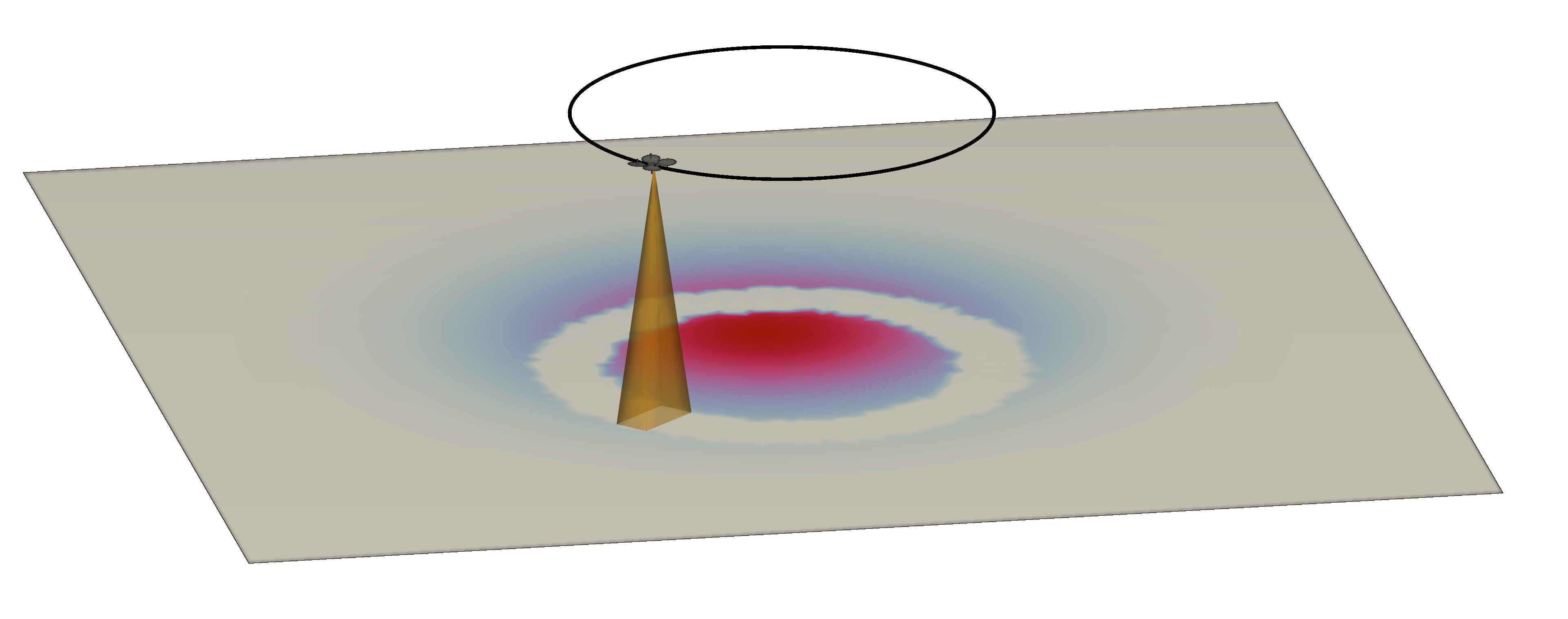}
	\caption{UAV stuck in a circular motion around the area of interest due to improperly defined parameters. To address this problem, an appropriate combination of minimal tuning radius, minimal altitude and the camera sensor (with adequate FOV) should be employed.} 
	\label{fig:small_fov}
\end{figure}

The proposed control can produce inadequate results if set-up inappropriately. One possible cause could be a lack of proper integration between the minimal turning radius, minimal altitude and the camera sensor, particularly its field of view. This error could result in an endless circular motion around the area of interest as shown in Figure \ref{fig:small_fov}. To prevent this issue, it is essential for the camera sensors' lateral scope to be greater than twice the minimal turning radius.

\begin{table}[!htb]
	\scriptsize
	\centering
	\caption{Incline compatibility confirmation for employed UAVs and test cases}
	\begin{tabularx}{\linewidth}{L{0.7}L{0.8}R{1.50}} 
		UAV type && Maximum supported terrain incline $\kappa$ \\
		\hline
		UAV A &&  76.9 $^\circ$ \\
		UAV B && 76.9 $^\circ$ \\
		UAV C && 59 $^\circ$	\\
		\\
		Test case & Utilized UAV types & Maximum terrain incline $\kappa_{T, max}$ \\
		\hline
		Plastic world & UAV A & 68.8 $^\circ$	\\
		Mt. Vesuvius & UAV A \& B & 75.8 $^\circ$\\
		Star dunes & UAV C & 54.1 $^\circ$ \\
	\end{tabularx}
	\label{tab:incline}
\end{table}

\begin{figure}[!ht]
	\centering
	\includegraphics[width=1\linewidth]{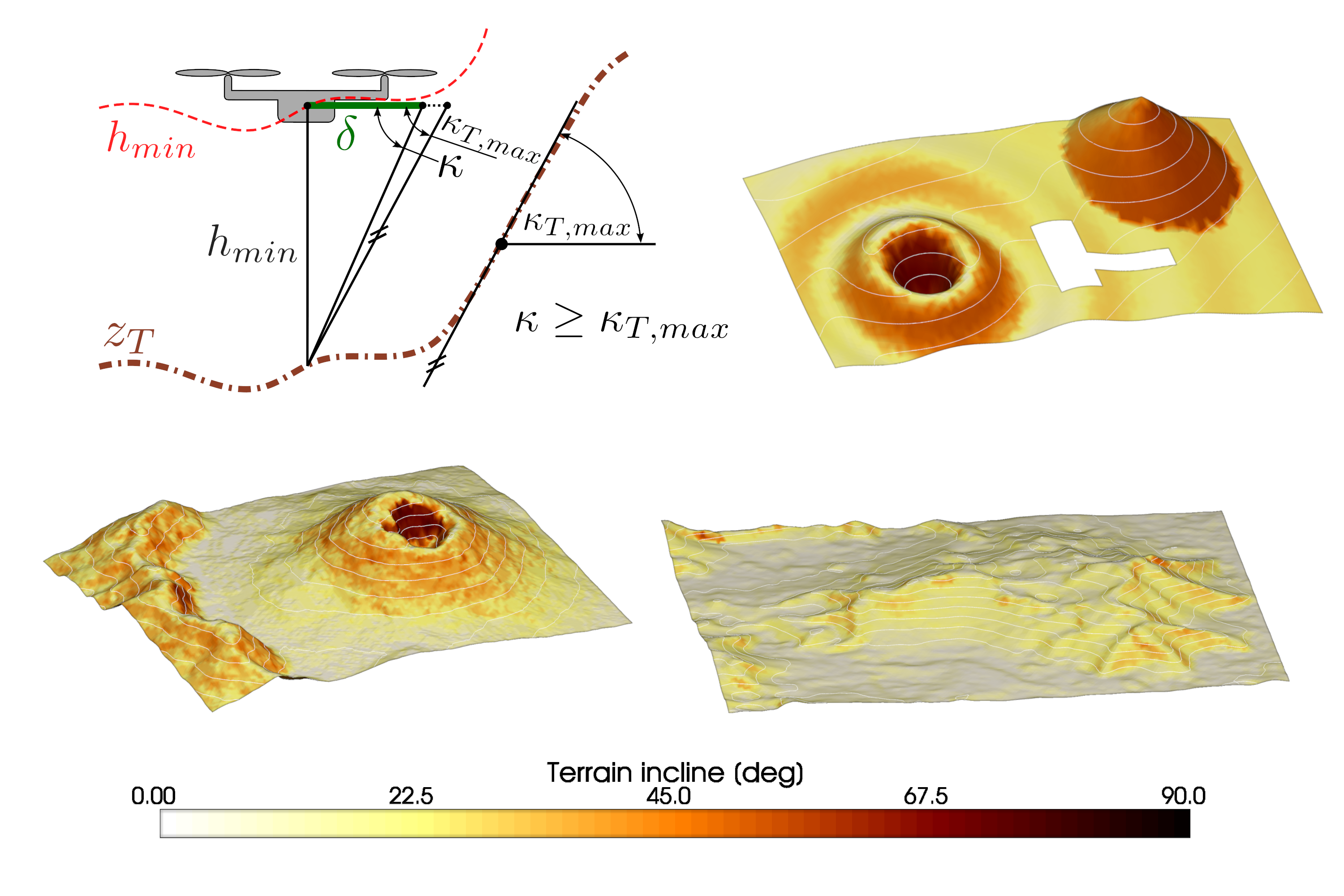}
	\caption{Sketch illustrating the idea behind UAVs' maximal supported incline calculation and terrain incline visualization for all used test cases cases.} 
	\label{fig:terrain_incline}
\end{figure}

The ability of navigating particular terrain with a designated UAV setup, considering clearance and minimal altitude values, relies on the maximum terrain incline. Due to the computational constraints associated with real-time control, direct horizontal clearance checks with the terrain are impractical within this algorithm. Instead, the algorithm determines if horizontal clearance is met across the complete domain while employing the specified minimal altitude. The worst case scenario is when the UAV is at its lowest possible altitude and the smallest possible horizontal clearance is attained near the terrain of maximal incline as shown in Figure \ref{fig:terrain_incline}. Therefore, for each UAV, we can calculate the maximal supported terrain incline   

\begin{equation*}
	\kappa_{i} = \arctan \left( \frac{h_{min,i}}{\delta_{i}} \right)
\end{equation*}

which has to be greater than maximal terrain incline $\kappa_{T, max}$. For our test cases we checked compatibility between the UAV fleet used and the domain as shown in Table \ref{tab:incline}. Terrain incline distribution across test cases is also shown in Figure \ref{fig:terrain_incline}. 

To explore a domain with an incline higher than supported while maintaining the desired minimal clearance, the only option (that does not include terrain modification/smoothing) is to raise the minimal flight altitude. However, for real-world implementation, this problem can be completely resolved by utilizing horizontal distance sensors usually integrated within the modern UAVs. 
\section{Conclusion}

We presented a centralized multi-UAV control algorithm that conducts complex terrain inspection and manages UAVs' flight in three-dimensional space. It concurrently addresses two-dimensional coverage problem, utilizing the Heat Equation Driven Area Coverage (HEDAC) method, and employs Model Predictive Control (MPC) for altitude and velocity regulation. Horizontal movement is driven by a potential field that is dynamically adjusted using custom sensory framework featuring a simulated real-time image capture and detection system. Implemented collision avoidance technique ensures adherence to clearance distance restrictions among multiple UAVs and with the boundary, while also satisfying the minimal turning radius constraints. It can be effectively employed with both multi-rotor and fixed-wing UAVs, depending on inspection requirements and the search domain characteristics.

The proposed method was employed on three test cases. The first test case consisted of challenging, synthetically generated terrain that was inspected through the collaborative action of 3 multi-rotor UAVs. The second case was an inspection of Mount Vesuvius that was conducted using 5 multi-rotor UAVs utilizing two different sensing setups with accompanying flight altitudes. The third case showcased a survey of desert dunes that was performed using two fixed-wing UAVs. Interesting sections of UAVs' trajectories were highlighted and thoroughly analyzed. In all three cases, the chosen parameters yielded satisfactory survey results with minimal computation time, facilitating real-time UAV control. The search success is the overall and final grade of the UAV control. It is evaluated with survey accomplishment $\eta$ metric in simulations of UAV search in realistic scenarios where every test case exceeded 80\%. That would translate into less than 20\% chance of an undetected target, on average, across the entire search domain.
The opportunity to increase $\eta$ exists through improvements in sensing/detection capabilities, UAV maneuverability, and reliable terrain data, along with an extension of flight time. Deploying more UAVs would undoubtedly have the most significant impact. With the scalability of HEDAC algorithm \cite{ivic2020motion} and the results presented here, authors are confident that additional improvements will allow a greater number of UAVs to be utilized in the search control.

The primary limitation of the method is its inability to navigate domains featuring significant terrain inclines without domain modification. However, in a practical application this problem can be resolved by employing proximity sensors usually installed in modern UAVs. 

In terms of future improvements, there are several enhancements that could be incorporated into the method. One possible enhancement involves implementing automatic terrain smoothing when dealing with steeply inclined domains. While implementing absolute distance checks between the UAV and the nearest terrain point sounds like an ideal scenario, it's crucial to acknowledge that this could result in a substantial increase in computation time, potentially making the method impractical for real-time UAV control. Another area for improvement is introducing the capability to selectively include or exclude UAVs from the search. This feature would address scenarios such as battery replacement, allowing for extended search duration within a single run. 

Furthermore, there are plans for future experimental validation of the proposed method. It includes implementing trajectory correction using GPS data which is essential to compensate for positioning errors caused by unpredictable external conditions and limitations of the UAV flight controller. Additionally, integration of the UAVs' proximity sensor would serve as a convenient way to address the horizontal clearance problem.

\section*{Acknowledgments}

This publication is supported by the Croatian Science Foundation under the project UIP-2020-02-5090.


\section*{Supplementary data}
All parameters for reproducing the study are presented in the manuscript. The data needed to reproduce the presented UAV search scenarios and video animations are available on the Open Science Framework repository: \url{https://osf.io/t947u/}. The Python code needed to reproduce this research is available upon request.

%
\IEEEpeerreviewmaketitle

\ifCLASSOPTIONcaptionsoff
  \newpage
\fi




\bibliographystyle{plain}
\bibliography{bibliography}




\end{document}